\documentclass[journal]{IEEEtran}

\usepackage{amssymb}
\usepackage{amsmath}
\usepackage{graphicx}
\usepackage{tabularx}
\usepackage{lipsum}
\usepackage[table,pdftex]{xcolor}
\usepackage{subfigure}
\usepackage{xspace}
\usepackage[colorlinks=true,citecolor=blue,urlcolor=black]{hyperref}
\usepackage[colorinlistoftodos]{todonotes}
\usepackage{bm}

\footernote{Accepted article to appear in IEEE Transactions on Medical Imaging, 2017. Please cite the journal version.}

\newcommand{\eqcm}{\textrm{ ,}}

\newcommand{\quott}[1]{`#1'}

\renewcommand{\vec}[1]{\bm{#1}}

\begin{document}

\title{Reverse Classification Accuracy:\\Predicting Segmentation Performance in the Absence of Ground Truth}

\author{Vanya~V.~Valindria, Ioannis~Lavdas, Wenjia~Bai, Konstantinos~Kamnitsas,\\Eric~O.~Aboagye, Andrea~G.~Rockall, Daniel~Rueckert, and Ben~Glocker 
\thanks{
V. Valindria, W. Bai, K. Kamnitsas, D. Rueckert, and B. Glocker are with the BioMedIA Group, Department of Computing, Imperial College London, SW7 2AZ, UK. E-mail: v.valindria15@imperial.ac.uk.}
\thanks{I. Lavdas and E.O. Aboagye are with the Comprehensive Cancer Imaging Centre, Hammersmith Hospital, Imperial College London, W12 0NN, UK}
\thanks{A.G Rockall is with The Royal Marsden NHS Foundation Trust, London, SW3 6JJ, UK}}

\maketitle

\begin{abstract}

When integrating computational tools such as automatic segmentation into clinical practice, it is of utmost importance to be able to assess the level of accuracy on new data, and in particular, to detect when an automatic method fails. However, this is difficult to achieve due to absence of ground truth. Segmentation accuracy on clinical data might be different from what is found through cross-validation because validation data is often used during incremental method development, which can lead to overfitting and unrealistic performance expectations. Before deployment, performance is quantified using different metrics, for which the predicted segmentation is compared to a reference segmentation, often obtained manually by an expert. But little is known about the real performance after deployment when a reference is unavailable. In this paper, we introduce the concept of \emph{reverse classification accuracy} (RCA) as a framework for predicting the performance of a segmentation method on new data. In RCA we take the predicted segmentation from a new image to train a reverse classifier which is evaluated on a set of reference images with available ground truth. The hypothesis is that if the predicted segmentation is of good quality, then the reverse classifier will perform well on at least some of the reference images. We validate our approach on multi-organ segmentation with different classifiers and segmentation methods. Our results indicate that it is indeed possible to predict the quality of individual segmentations, in the absence of ground truth. Thus, RCA is ideal for integration into automatic processing pipelines in clinical routine and as part of large-scale image analysis studies.

\end{abstract}


%
\IEEEpeerreviewmaketitle

\section{Introduction}

Segmentation is an essential component in many image analysis pipelines that aim to extract clinically useful information from medical images to inform clinical decisions in diagnosis, treatment planning, or monitoring of disease progression. A multitude of approaches have been proposed for solving segmentation problems, with popular techniques based on graph cuts \cite{boykov2006graph}, multi-atlas label propagation \cite{iglesias2015multi}, statistical models \cite{heimann2009statistical}, and supervised classification \cite{geremia2013classification}. Traditionally, performance of a segmentation method is evaluated on an annotated database using various evaluation metrics in a cross-validation setting. These metrics reflect the performance in terms of agreement \cite{zou2004statistical} of a predicted segmentation compared to a reference \quott{ground truth} (GT)\footnote{For simplicity, we use the term ground truth to refer to the best known reference, which is typically a manual expert segmentation.}. Commonly used metrics include Dice's similarity coefficient (DSC) \cite{dice1945measures} and other overlap based measures \cite{crum2006generalized}, but also metrics based on volume differences, surface distances, and others \cite{taha2015metrics,deng2007simulating,ledig2014patch}. A detailed analysis of common metrics and their suitability for segmentation evaluation can be found in \cite{konukoglu2012discriminative}.

Once a segmentation method is deployed in routine little is known about its real performance on new data. Due the absence of GT, it is not possible to assess performance using traditional evaluation measures. However, it is critical to be able to assess the level of accuracy on clinical data \cite{van20073d}, and in particular, it is important to detect when an automatic segmentation method fails. Especially when the segmentation is an intermediate step within a larger automated processing pipeline where no visual quality control of the segmentation results is feasible. This is of high importance in large-scale studies such as the UK Biobank Imaging Study \cite{sudlow2015uk} where automated methods are applied to large cohorts of several thousand images, and the segmentation is to be used for further statistical population analysis. In this work, we are asking the question whether it is possible to assess segmentation performance and detect failure cases when there is no GT available to compare with. One possible approach to monitor the segmentation performance is to occasionally select a random dataset, obtain a manual expert segmentation and compare it to the automatic one. While this can merely provide a rough estimate about the average performance of the employed segmentation method, in clinical routine we are interested in the per case performance and want to detect when the automated method fails. The problem is that the performance of a method might be substantially different on clinical data, and is usually lower than what is found through cross-validation on annotated data carried out beforehand due to several reasons. First, the annotated database is normally used during incremental method development for training, model selection and fine tuning of hyper-parameters. This can lead to overfitting \cite{dwork2015reusable} which is a potential cause for lower performance on new data. Second, the clinical data might be different due to varying imaging protocols or artefacts caused by pathology. To this end, we propose a general framework for predicting the real performance of deployed segmentation methods on a per case basis in the absence of GT.

\subsection{Related Work}

Retrieving an objective performance evaluation without GT has been an issue in many domains, from remote sensing \cite{baraldi2005quality}, graphics \cite{liu2013no}, to marketing strategies \cite{cerrato2011classification}. In computer vision, several works evaluate the segmentation performance by looking at contextual properties \cite{correia2002stand}, by separating the perceptual salient structures \cite{ge2007new}, or by automatically generating semantic GT \cite{goldmann2008towards, li2013benchmark}. However, these methods cannot be applied to a more general task, such as an image with many different class labels to be segmented. An attempt to compute objective metrics, such as precision and recall with missing GT is proposed by \cite{lamiroy2013computing} but it cannot be used for data sets with partial GT since it applies a probabilistic model under the same assumptions. Another stand-alone method is to consider a meta-evaluation framework, where image features are used in a machine learning setting to provide a ranking of different methods \cite{zhang2006meta}. But this does not allow the estimation of segmentation performance on an individual image level.

Meanwhile, unsupervised methods \cite{chabrier2006unsupervised,zhang2008image} aim to estimate the segmentation accuracy directly from the images and labelmaps using, for example, information-theoretic and geometrical features. While unsupervised methods can be applied to scenarios where the main purpose of segmentation is to yield visually consistent results that are meaningful to a human observer, the application in medical settings is unclear. 

When there are multiple reference segmentations available, a similarity measure index can be obtained by comparing an automatic segmentation with the set of references \cite{unnikrishnan2007toward}. In medical imaging, the problem of performance analysis with multiple references which may suffer from intra-rater and inter-rater variability, has been addressed \cite{warfield2004simultaneous, li2010estimating}. The STAPLE approach \cite{warfield2004simultaneous} has lead to the work of Bouix et al. \cite{bouix2007evaluating} that proposed techniques for comparing the relative performance of different methods without the need of GT. Here, the different segmentation results are treated as plausible references, thus can be evaluated through STAPLE and the concept of common agreement. Another work by \cite{sikka2010comparison} has quantitatively evaluated the performance of several segmentation algorithms by region-correlation matrix. The limitation of this work is that it cannot evaluate the segmentation performance of a particular method on a particular image independently.

Recent work has explored the idea of learning a regressor to directly predict segmentation accuracy from a set of features that are related to various segmentation energy terms \cite{kohlberger2012evaluating}. Here, the assumption is that those features are well suited to characterise segmentation quality. In an extension for a security application, the same features as in \cite{kohlberger2012evaluating} are extracted and used to learn a generative model of good segmentations that can be used to detect outliers \cite{grady2012automatic}. Similarly, the work of \cite{frounchi2011automating} considers training of a classifier that is able to discriminate between consistent and inconsistent segmentations. However, the approaches \cite{kohlberger2012evaluating, frounchi2011automating} can only be applied when a training database with good and bad segmentations is available from which a mapping from features to segmentation accuracy can be learned. Examples of bad segmentations can be generated by altering parameters of automatic methods, but it is unclear whether those examples resemble realistic cases of segmentation failure. The generative model approach in \cite{grady2012automatic} is appealing as it only requires a database of good segmentations. However, there is still the difficulty of choosing appropriate thresholds on the probabilities that indicate bad or failed segmentations. Such an approach cannot not be used to directly predict segmentation scores such as DSC, but can be useful to inform automatic quality control or to automatically select the best segmentation from a set of candidates.

In the general machine learning domain, the lack-of-label problem has been tackled by exploiting transfer learning \cite{zhong2010cross} using a reverse validation to perform cross-validation when the number of labeled data is limited. The basic idea of reverse validation \cite{zhong2010cross} is based on reverse testing \cite{fan2006reverse}, where a new classifier is trained on predictions on the test data and evaluated again on the training data. This idea of reverse testing is closely related to our approach as we will discuss in the following.

\subsection{Contribution}

The main contribution of this paper is the introduction of the concept of \emph{reverse classification accuracy} (RCA) to assess the segmentation quality of an individual image in the absence of GT. RCA can be applied to evaluate the performance of any segmentation method on a per case basis. To this end, a classifier is trained using a single image with its predicted segmentation acting as \emph{pseudo} GT. The resulting \emph{reverse classifier} (or RCA classifier) is then evaluated on images from a reference database for which GT is available. It should be noted that the reference database can be (but does not have to be) the training database that has been used to train, cross-validate and fine-tune the original segmentation method. The assumption is that in machine learning approaches, such a database is usually already available, but it could also be specifically constructed for the purpose of RCA. Our hypothesis is that if the segmentation quality for a new image is high, then the RCA classifier trained on the predicted segmentation used as pseudo GT will perform well at least on some of the images in the reference database, and similarly, if the segmentation quality is poor, the classifier is likely to perform poorly on the reference images. For the segmentations obtained on the reference images through the RCA classifier, we can quantify the accuracy, e.g., using DSC, since reference GT is available. It is expected that the maximum DSC score over all reference images correlates well with the real DSC that one would get on the new image if GT were available.
While the idea of RCA is similar to reverse validation \cite{zhong2010cross} and reverse testing \cite{fan2006reverse}, the important difference is that in our approach we train a reverse classifier on every single instance while the approaches in \cite{zhong2010cross, fan2006reverse} train single classifiers over the whole test set and its predictions jointly to find out what the best original predictor is. RCA has the advantage of allowing to predict the accuracy for each individual case, while at the same time aggregating over such accuracy predictions allows drawing conclusions for the overall performance of a particular segmentation method.

In the following, we will first present the details of RCA and then evaluate its applicability to a multi-organ segmentation task by exploring the prediction quality of different segmentation metrics for different combinations of segmentation methods and RCA classifiers. Our results indicate that, at least to some extent, it is indeed possible to predict the performance level of a segmentation method on each individual case, in the absence of ground truth. Thus, RCA is ideal for integration into automatic processing pipelines in clinical routine and as part of large-scale image analysis studies.
                      
\section{Reverse Classification Accuracy}

\begin{figure}
\centering
  \includegraphics[width=0.97\linewidth]{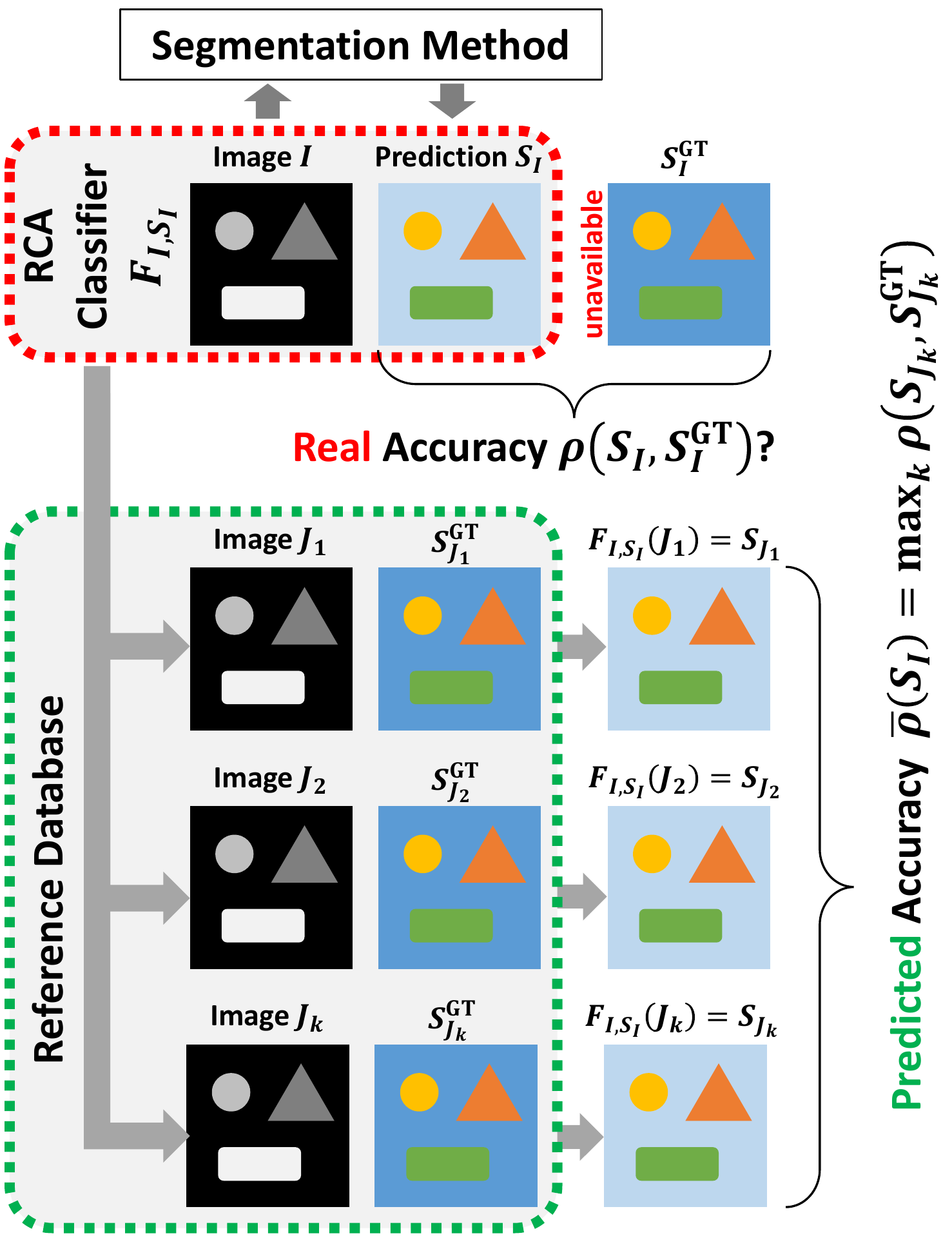}
\caption{Overview of the general framework for reverse classification accuracy. The accuracy of the predicted segmentation $S_I$ of a new image $I$ is estimated via a RCA classifier trained using $S_I$ as pseudo ground truth, and applied to a set of reference images $J_k$ for which ground truth reference segmentations $S_{J_k}^{\textrm{GT}}$ are available. The best segmentation score computed on the reference images is used as prediction of the real accuracy of segmentation $S_I$.}\label{fig:overview}
\end{figure}

The RCA framework is based on the idea of training reverse classifiers on individual images utilizing their predicted segmentation as pseudo GT. An overview of the RCA framework is shown in Fig. \ref{fig:overview}. In this work, we employ three different methods for realizing the RCA classifier and evaluate each in different combinations with three state-of-the-art image segmentation methods. Details about the different RCA classifiers are provided in the following.

\subsection{Learning Reverse Classifiers}

Given an image $I$ and its predicted segmentation $S_I$, we aim to learn a function $f_{I,S_I}(\vec{x})\!:\! \mathbb{R}^n \mapsto C$ that acts as a classifier by mapping feature vectors $\vec{x} \in \mathbb{R}^n$ extracted for individual image points to class labels $c \in C$. In theory, any classification approach could be utilized within the RCA framework for learning the function $f_{I,S_I}$. 
We experiment with three different methods reflecting state-of-the-art machine learning approaches for voxel-wise classification and atlas-based label propagation.

\paragraph{Atlas Forests} The first approach we consider for learning a RCA classifier is based on the recently introduced concept of Atlas Forests (AFs) \cite{zikic2014encoding} which demonstrates the feasibility of using Random Forests (RFs) \cite{breiman2001random} to encode individual atlases, i.e., images with corresponding segmentations. Random Forests have become popular for general image segmentation tasks as they naturally handle multi-class problems and are computationally efficient. Since they operate as voxel-wise classifiers, they do not (necessarily) require pre-registration of the images neither at training nor testing time. Although in \cite{zikic2014encoding} spatial priors have been incorporated by means of registering location probability maps to each atlas and new image, this is not a general requirement for using AFs to encode atlases. In fact, the way we employ AFs within our RCA framework does not require any image registration. The forest-based RCA classifiers in this work are trained all with the same set of parameters of maximum depth 30 and 50 trees. As we follow a very standard approach for RFs, we refer to \cite{criminisi2012decision,zikic2014encoding} for more details. Worth to note that, similar to previous work, we employ simple box features which can be efficiently evaluated using integral images. This has the advantage that feature responses do not need to be precomputed. Instead, we randomly generate a large pool of potential features (typically around 10,000) by drawing randomly values for the feature parameters such as box sizes and offsets from predefined ranges. At each split node we then evaluate on-the-fly a few hundred box features with a brute force search for optimal thresholds over the range of feature responses to greedily find the most discriminative feature/threshold pair. This strategy has proven successful in a number of works using RFs for various tasks.

\paragraph{Deep Learning} We also experiment with convolutional neural networks (CNNs) as RCA classifiers. Here, we utilize DeepMedic\footnote{Code available at \url{https://github.com/Kamnitsask/deepmedic}}, a 3D CNN architecture for automatic segmentation \cite{kamnitsas2016efficient}. The architecture is computationally efficient as it can handle large image context by using a dual pathway for multi-scale processing. CNNs have been shown to be able to learn highly complex and discriminative data associations between input data and target output. The architecture of the network is defined by the number of layers and the number of activation functions in each layer. In CNNs, each activation function corresponds to a learned convolutional filter, and each filter produces a feature map (FM) by convolving the outputs of the previous layer. Through the sequential application of many convolutions, highly complex features are learned that are then used to produce voxel-wise predictions at the final, fully-connected layer. CNNs are a type of deep learning approach which normally requires large amounts of training data in order to perform well due to the thousands (or millions) of parameters corresponding to the weights of the filters. To be able to act as a RCA classifier that is trained on a single image, we require a specialised architecture. Here, we reduce the number of FMs in each layer by one third compared to the default setting of DeepMedic. We also cut the feature maps in the last fully connected layers, from 150 to 45. By reducing the feature maps without changing the architecture in terms of number of layers the network preserves its capability to see large image context as the size of the receptive field remains unchanged. With less number of filters, the number of parameters is substantially decreased, which leads to faster computations, but more importantly, reduces overfitting when trained on a single image. Training is performed in a patch-wise manner where the original input image is devided into 3D patches that are then sampled during training using backpropagation and batch normalization. For details about the training procedure and further analysis of DeepMedic we refer to \cite{kamnitsas2016efficient}.

\paragraph{Atlas-based Label Propagation} The third approach we consider is atlas-based label propagation. Label propagation using multiple atlases have been shown to yield state-of-the-art results on many segmentation problems \cite{iglesias2015multi}. A common procedure in multi-atlas methods is to use non-rigid registration to align the atlases with the image to be segmented and then perform label fusion strategies to obtain predictions for each image point. Although, multi-atlas methods based on registration are not strictly voxel-wise classifiers as they operate on the whole image during registration, the final stage of label fusion can be considered as a voxel-wise classification step. Here, we make use of an approach that has been originally developed in the context of segmentation of cardiac MRI \cite{bai2013probabilistic}\footnote{Code available at \url{https://github.com/baiwenjia/CIMAS}}. For the purpose of RCA, however, there is only a single atlas and thus no label fusion is required. Using single atlas label propagation then boils down to making use of an efficient non-rigid registration technique as the one described in \cite{bai2013probabilistic}. For RCA, the single atlas then corresponds to the image and its predicted segmentation for which we want to estimate the segmentation quality. We use the same configuration for image registration as in \cite{bai2013probabilistic} and refer to this work for further details.

\subsection{Predicting Segmentation Accuracy}

For the purpose of assessing the quality of an individual segmentation, we train a RCA classifier $f_{I,S_I}$ on a single image $I$ that has been segmented by any segmentation method, where $S_I$ denotes the predicted segmentation that here acts as pseudo GT during classifier training. Our objective is to estimate the quality of $S_I$ in the absence of GT. To this end, we define the segmentation function $F_{I,S_I}(J)=S_J$ that applies the trained RCA classifier $f_{I,S_I}$ to all voxels (or more precisely to the features extracted at each voxel) of another image $J$ which produces a segmentation $S_J$. Assuming that for the image $J$ a reference GT segmentation $S_J^{\textrm{GT}}$ is available, we can now compute any segmentation evaluation metric on the pair $(S_J,S_J^{\textrm{GT}})$. The underlying hypothesis in our RCA framework is that there is a correlation between the values computed on $(S_J,S_J^{\textrm{GT}})$ and the values one would get for the pair $(S_I,S_I^{\textrm{GT}})$, where $S_I^{\textrm{GT}}$ is the reference GT of image $I$ which in practice, however, is unavailable.

It is unlikely that this assumption of correlation holds for an arbitrary reference image $J$. In fact, the RCA classifier $f_{I,S_I}$ is assumed to work best on images that are somewhat similar to $I$. Therefore, we further assume that a suitable reference database is available that contains multiple segmented images (or atlases) ${\cal{T}}=\{(J_k,S_{J_k}^{\textrm{GT}})\}_{k=1}^m$ that capture the expected variability. Such a database is commonly available in the context of machine learning and multi-atlas based segmentation approaches, but could also be generated specifically for the purpose of RCA. If already available, we can re-use existing training databases that might have been previously used during method development and/or cross-validation and parameter tuning. When testing the RCA classifier on all of the available $m$ reference images, we expect that the RCA classifier performs well on at least some of these, if and only if the predicted segmentation $S_I$ is of good quality. If $S_I$ is of bad quality, we expect the RCA classifier to perform poorly on all reference images. This leads to our definition of a proxy measure for predicting the segmentation accuracy as
\begin{align}
\label{eq:rca}
\bar{\rho}(S_I) = \max_{1 \leq k \leq m} \rho(F_{I,S_I}(J_k), S_{J_k}^{\textrm{GT}}) \eqcm
\end{align}
where $\rho$ is any evaluation metric, such as DSC, assuming higher values correspond to higher quality segmentations\footnote{For metrics where a lower value indicates better quality, such as surface distance, we can simply replace the $\max$ with a $\min$ operator.}. Here, we only look for the maximum value that is found across all reference images, as this seems to be a good indicator of the quality of the segmentation $S_I$. Other statistics could be considered, such as the average of the top three scores, but we found that the maximum score works best as a proxy. Note, that the mean or median scores are not very useful measures as we do not expect the RCA classifier to work well on the majority of the reference images. Afterall, the RCA classifier does overfit to the single image and will not generalize to perform well on dissimilar images. Nonetheless, as we will demonstrate in the experiments, $\bar{\rho}$ indeed provides accurate estimates for the segmentation quality in a wide range settings

\subsection{Summary}
The following provides a summary of the required steps for using RCA in practice within a processing pipeline for automatic image segmentation. Given an image $I$ to be segmented:
\begin{enumerate}
\item Run the automated image segmentation method to obtain predicted segmentation $S_I$.
\item Train a RCA classifier on image $I$ and its predicted segmentation $S_I$ to obtain an image segmenter $F_{I,S_I}$. 
\item Evaluate the RCA classifier on a reference database with images for which GT is available ${\cal{T}}=\{(J_k,S_{J_k}^{\textrm{GT}})\}_{k=1}^m$ to obtain segmentations $\forall k \, F_{,S_I}(J_k) = S_{J_k}$.
\item Compute the segmentation quality of $S_I$ using a proxy measure $\bar{\rho}(S_I)$ according to Eq. (\ref{eq:rca}).
\end{enumerate}
Depending on the application, a threshold may be defined on $\bar{\rho}$ to flag up images with poor segmentation quality that need manual inspection, or to automatically identify high quality segmentations suitable for further analysis.

\section{Experimental Validation}
In order to test the effectiveness of the RCA framework, we explore a comprehensive multi-organ segmentation task on whole-body MRI. In this application, we evaluate the prediction accuracy of RCA in the context of three different state-of-the-art segmentation methods, a Random Forest approach \cite{geremia2013classification}, a deep learning approach using 3D CNNs \cite{kamnitsas2016efficient}, and a probabilistic multi-atlas label propagation approach \cite{bai2013probabilistic}. The dataset used to validate our framework is from our MALIBO (MAchine Learning In whole Body Oncology) study. We collected whole-body, multi-sequence MRI (T1w Dixon and T2w images) of 35 healthy volunteers. Detailed manual segmentations of 15 anatomical structures, including abdominal organs (heart, left/right lung, liver, adrenal gland, gall bladder, left/right kidney, spleen, pancreas, bladder) and bones (spine, left/right clavicle, pelvis) have been generated by clinical experts as part of the study. These manual segmentations will serve as GT in the quantitative evaluation.

\subsection{Experimental Setting}
We use 3-fold cross validation to automatically segment all 525 structures (15 organs $\times$ 35 subjects) with each of the three different segmentation methods, namely Random Forests, CNNs, and Multi-Atlas. In each fold, we use the RCA framework with three different methods for realizing the RCA classifier, namely Atlas Forests, constrained CNNs, and Single-Atlas, as described above. Using the RCA classifiers that are trained on each image for which we want to assess segmentation quality, we obtain segmentations on all reference images which are then compared to their manual reference GT. Since the GT is available for all 35 cases, we can compare the predicted versus the real segmentation accuracy for all cases and all organs under various settings with nine different combinations of segmentation methods and RCA classifiers.

\subsection{Quantifying Prediction Accuracy}
The Dice's similarity coefficient is the most widely used measure for evaluating segmentation performance\footnote{Despite some well known shortcomings of DSC as discussed in \cite{konukoglu2012discriminative}.}, and in our main results we focus on evaluating how well DSC can be predicted using our RCA framework. In order to quantify prediction accuracy, we consider three different measures, namely the correlation between predicted and real DSC, the mean absolute error (MAE), and a classification accuracy. Arguably, the most important measure for direct evaluation of how well RCA works is the MAE, as it directly tells us how close the predicted DSC is to the real one. Correlation is interesting, as it tells us something about the relation between predicted and real scores. We expect high correlation in order for RCA to be useful, but we might not always have an identity relation, as there could be a bias in the predictions. For example, if the predicted score is consistently lower than the real score, this can still be useful in practice, and will be indicated by high correlation but might not yield low MAEs. In such a case, a calibration might be considered as we will discuss later on. We also explore whether the predictions can be used to categorize segmentations according to their quality. We argue that for many clinical applications it is already of great value to be able to discriminate between good, bad, and possibly medium quality segmentations and that the absolute segmentation scores are of less importance. For proof-of-principle, we consider a three-category classification by grouping segmentations within DSC ranges $[0.0,0.6)$ for \quott{bad}, $[0.6,0.8)$ for \quott{medium}, and $[0.8,1.0]$ for \quott{good} cases. Note, that those ranges are somewhat arbitrary, in particular, as the quality of absolute DSC values is highly depending on the structure of interest. So in practice, those ranges would need to be adjusted specifically to the application at hand.

\begin{table*}[ht]
\centering
\caption{Predicting DSC for different segmentation methods using different RCA classifiers. Best results are obtained with the Single-Atlas approach yielding the lowest error between real and predicted DSC scores and high classification accuracy.}
\label{tab:main-results}
\begin{tabular}{ l  l | c c | c c | c c |}
& & \multicolumn{2}{c|}{Correlation} & \multicolumn{2}{c|}{MAE} & \multicolumn{2}{c|}{Accuracy 3-Categories} \\
Segmentation Method & RCA Classifier & All & No Zeros & All & No Zeros & All & No Zeros \\
\hline
Random Forests & Atlas Forests  & 0.881       & 0.867         & 0.120       & 0.130          & 0.783          & 0.776         \\
CNNs           & Atlas Forests  & 0.828       & 0.630          & 0.166      & 0.245         & 0.623          & 0.500          \\
Multi-Atlas    & Atlas Forests  & 0.863       & 0.877         & 0.168      & 0.177         & 0.749          & 0.726         \\
\hline
Random Forests & Constrained CNNs            & 0.721       & 0.718         & 0.252      & 0.271         & 0.653          & 0.631         \\
CNNs           & Constrained CNNs            & 0.756       & 0.662         & 0.225      & 0.292         & 0.592          & 0.472         \\
Multi-Atlas    & Constrained CNNs            & 0.773       & 0.686         & 0.209      & 0.237         & 0.693          & 0.642         \\
\hline
Random Forests & Single-Atlas    & 0.955       & 0.946         & 0.051      & 0.052         & 0.888          & 0.880          \\
CNNs           & Single-Atlas    & 0.973       & 0.892         & 0.052      & 0.065         & 0.811          & 0.756         \\
Multi-Atlas    & Single-Atlas    & 0.962       & 0.947         & 0.067      & 0.072         & 0.822          & 0.798         \\
       
\end{tabular}
\end{table*}
\begin{figure*}[t]
	\includegraphics[width=0.29\linewidth]{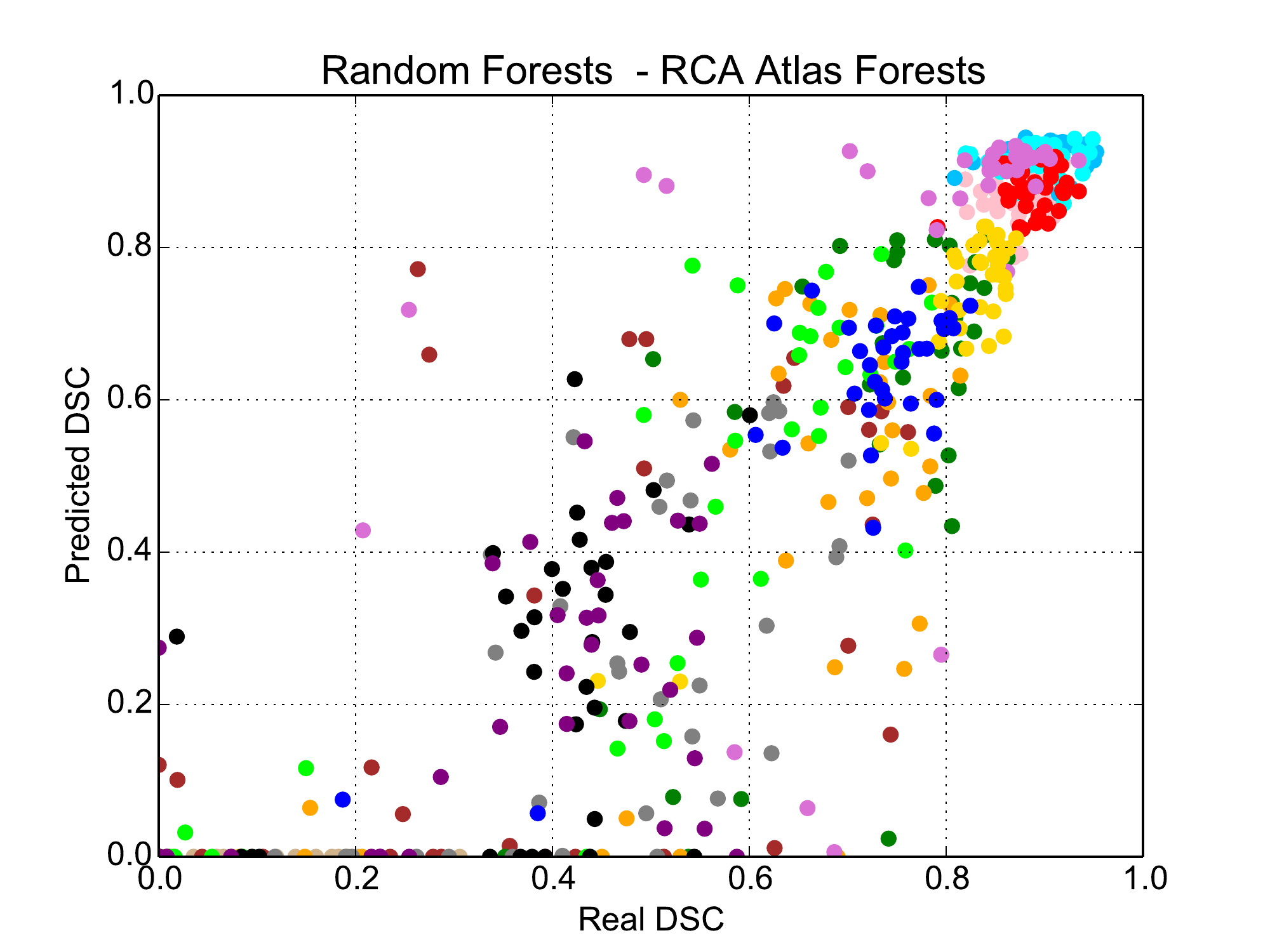}
	\includegraphics[width=0.29\linewidth]{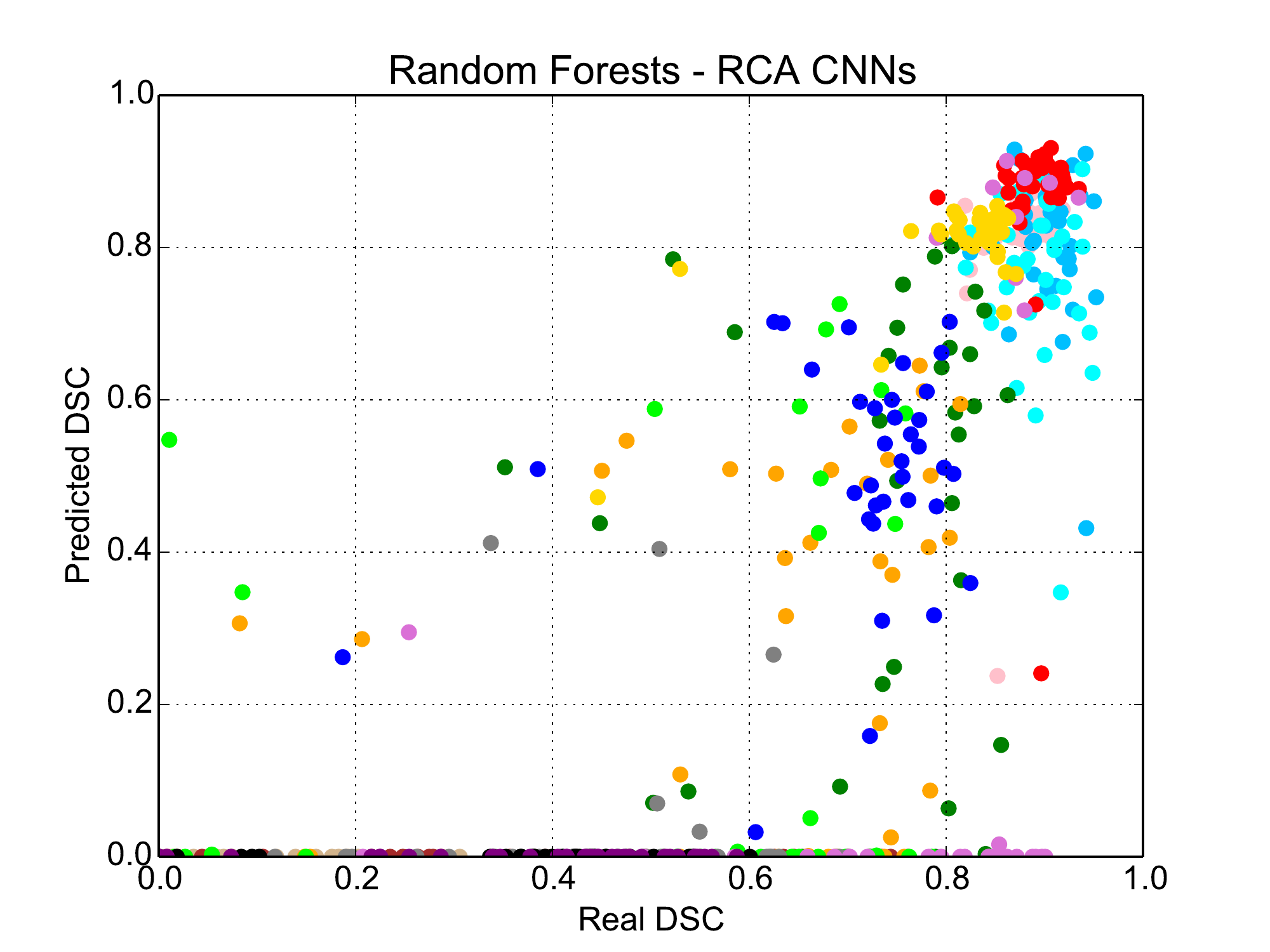}
	\includegraphics[width=0.29\linewidth]{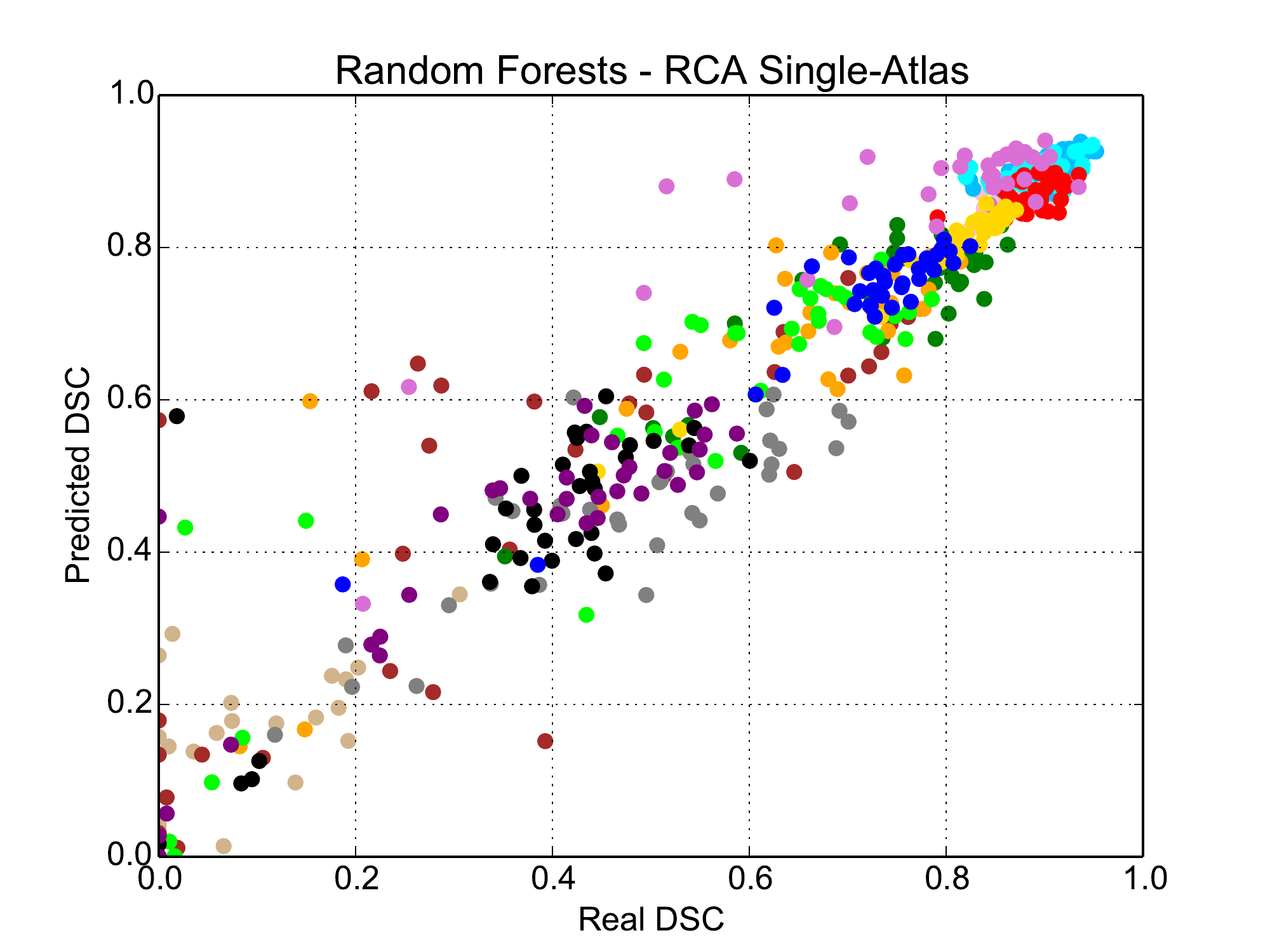} \\
	\includegraphics[width=0.29\linewidth]{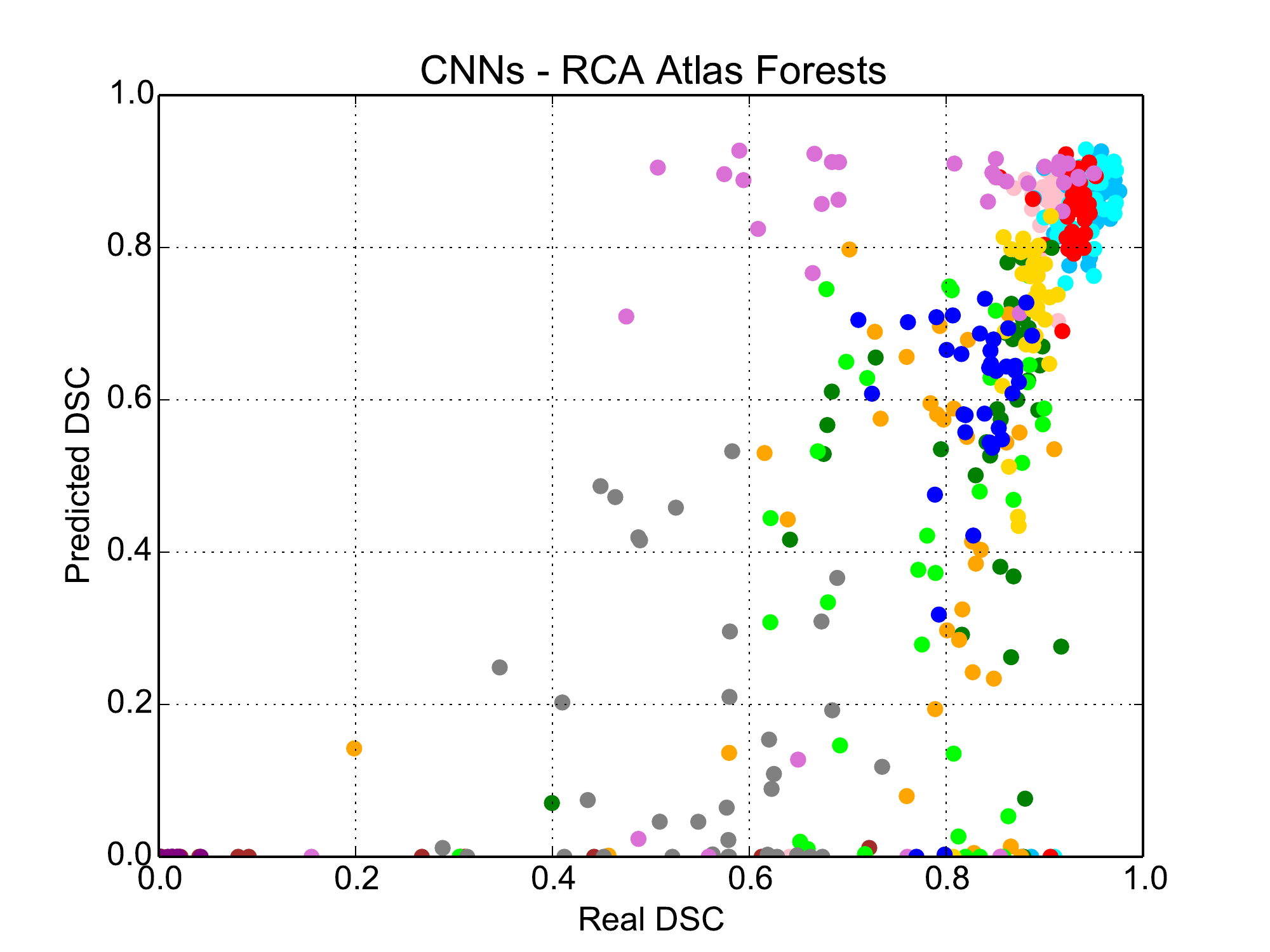}
	\includegraphics[width=0.29\linewidth]{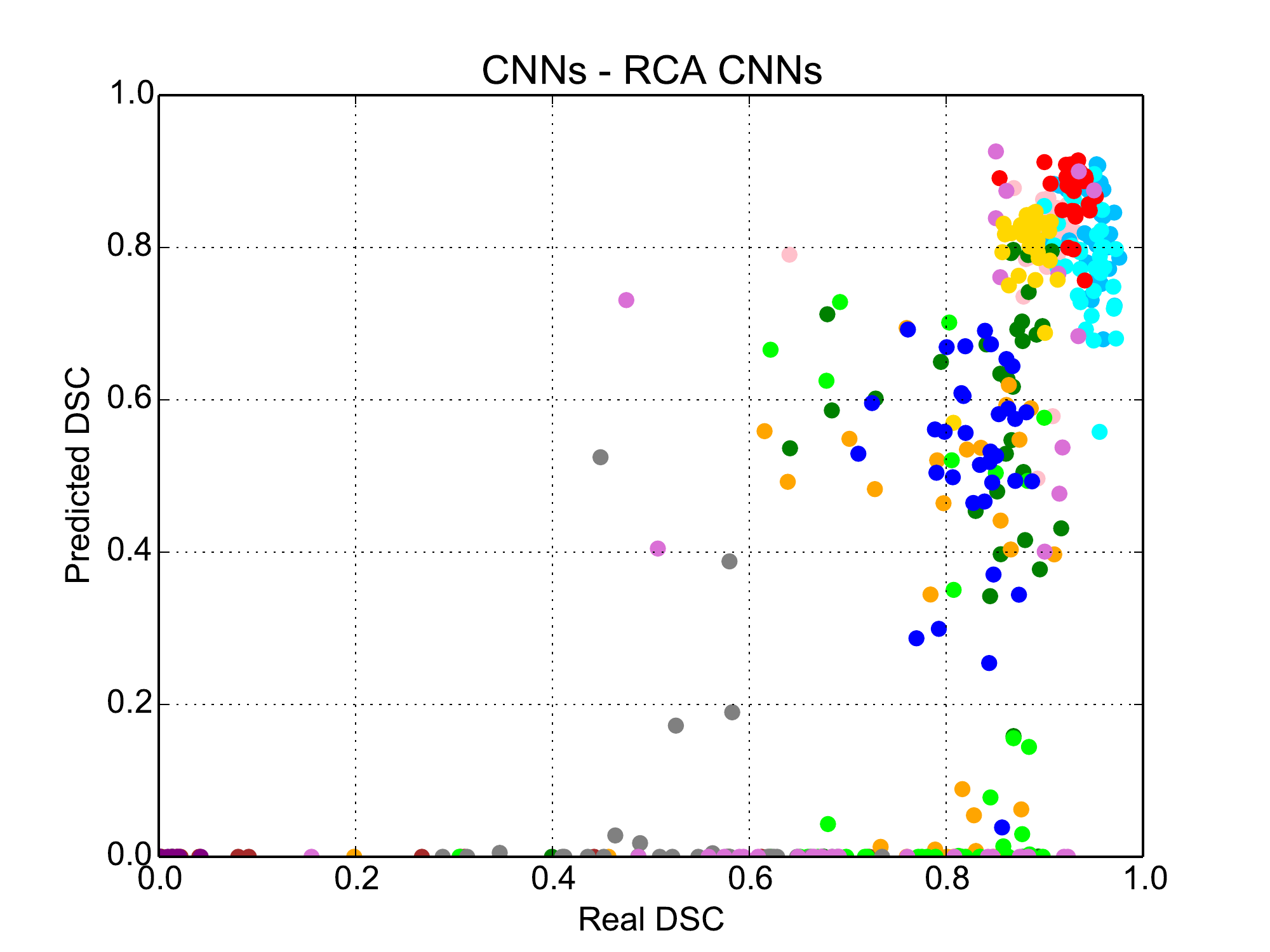}
	\includegraphics[width=0.29\linewidth]{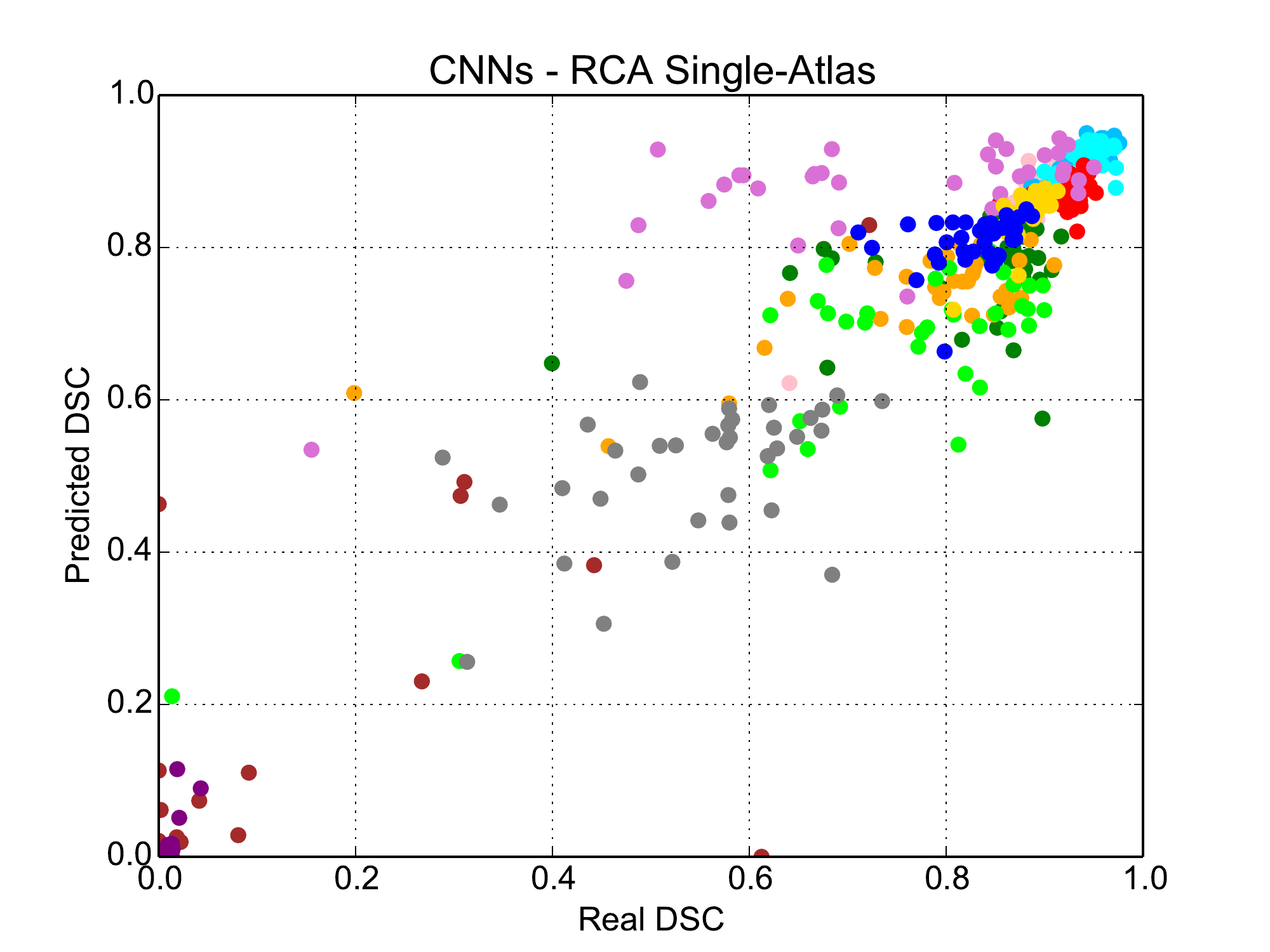}
	\includegraphics[width=0.10\linewidth]{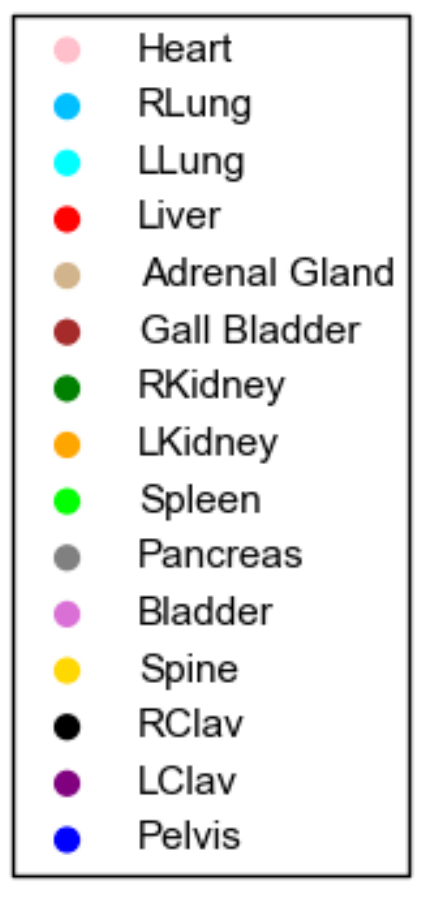} \\
	\includegraphics[width=0.29\linewidth]{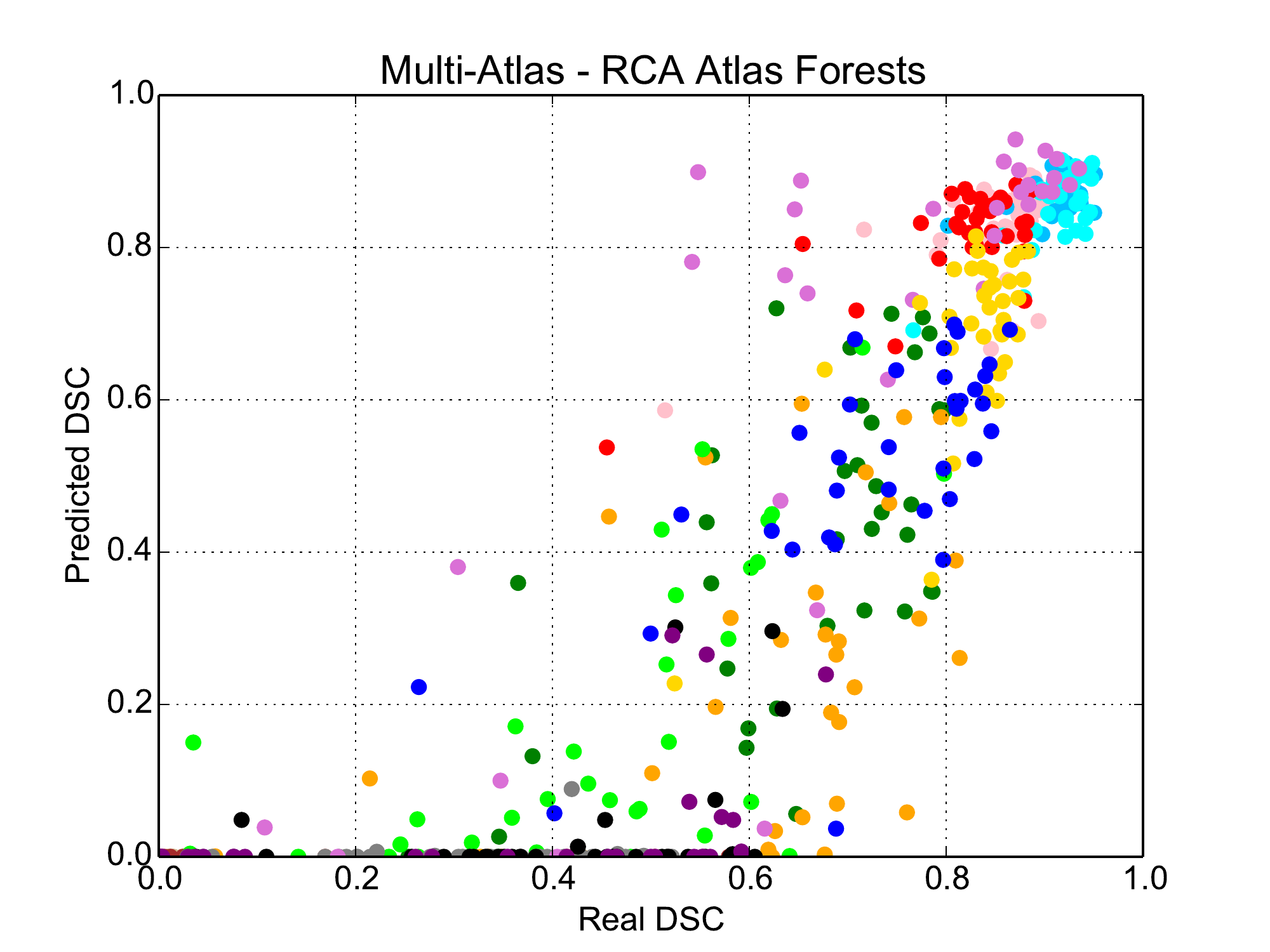}
	\includegraphics[width=0.29\linewidth]{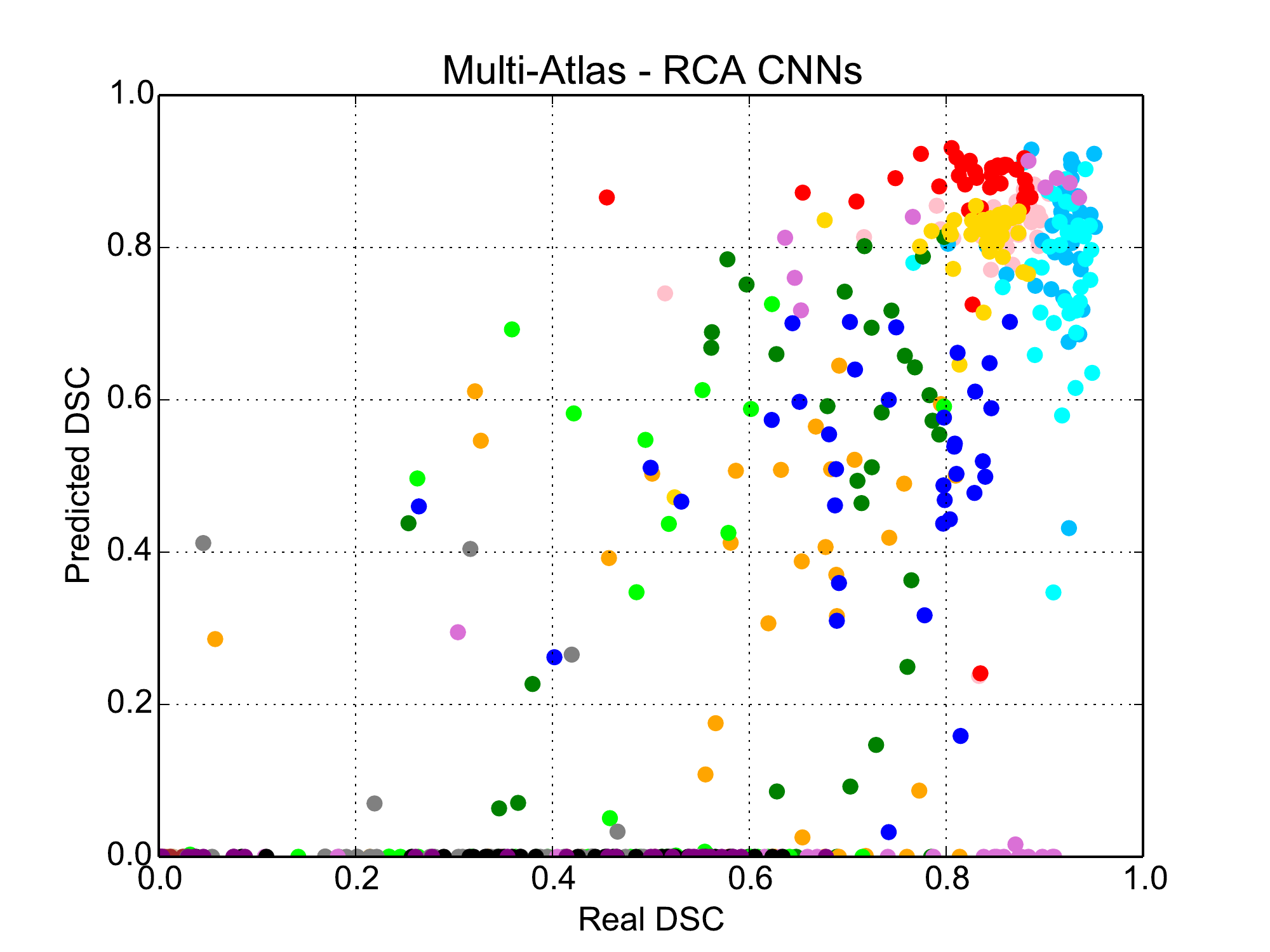}
	\includegraphics[width=0.29\linewidth]{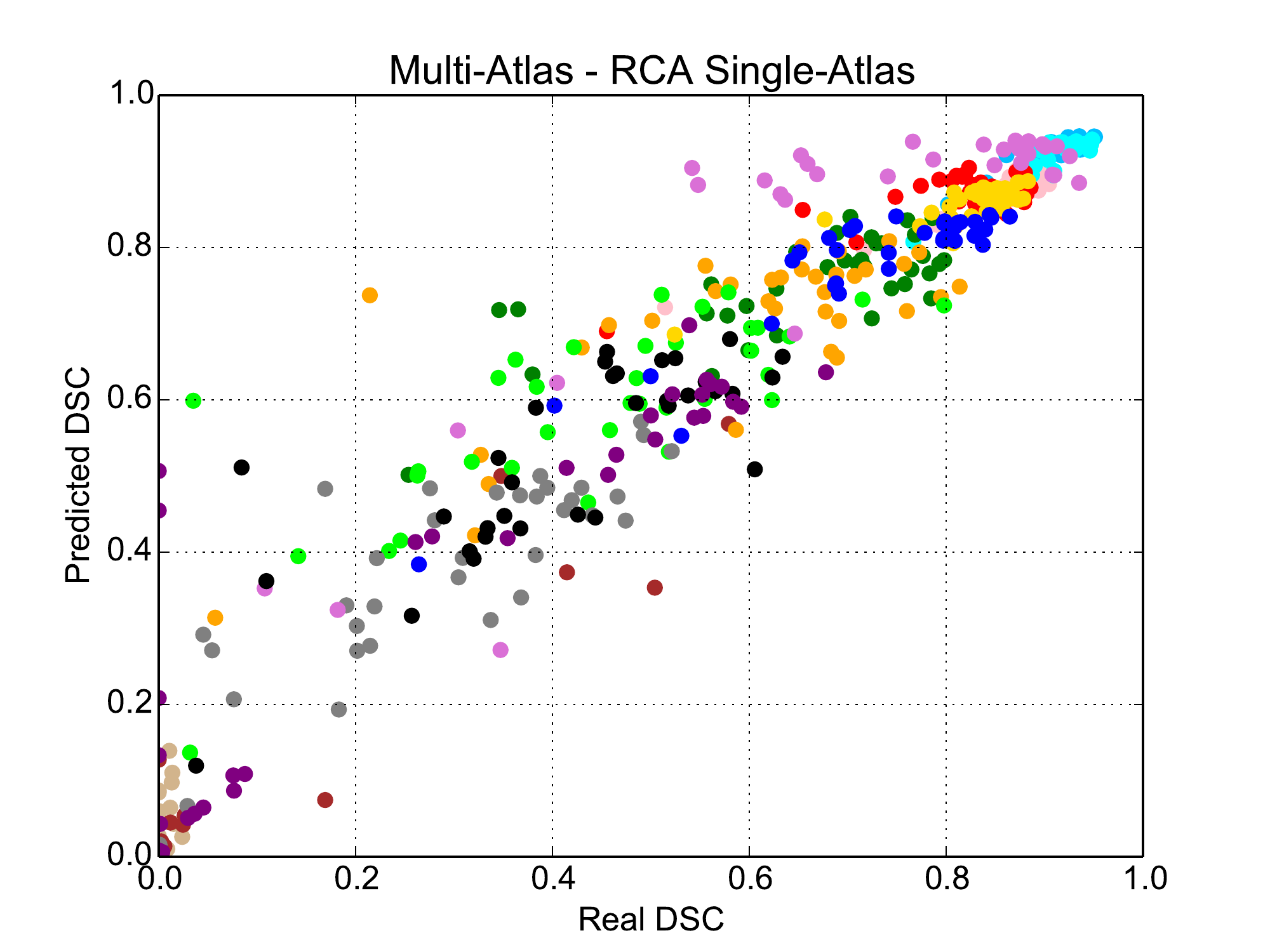}    
\caption{Scatter plots of predicted and real DSC of multiple structures for three different segmentation methods (rows) using three different RCA classifiers (columns). High correlation and low prediction errors are obtained when employing the single-atlas label propagation as RCA classifier (right column). There is also good correlation with predictions in case of Atlas Forests (left) with larger spread towards lower quality segmentations. The constrained CNNs (middle column) are less suitable for RCA which is likely due to the difficulty of training on single images. Both, Atlas Forests and constrained CNNs work best for larger organs such as liver, lungs, and pelvis while leading to many zero predictions for smaller structures such as adrenal gland and clavicles. The single-atlas label propagation makes accurate predictions of segmentation quality across all 15 anatomical structures. A summary of the plots is given in Tab.~\ref{tab:main-results}.}\label{fig:main-scatter}
\end{figure*}

\subsection{Results for Predicting Dice's Similarity Coefficients}
Our main results are summarized in Tab.~\ref{tab:main-results} where we report the quantitative analysis of the predicted accuracy for nine different settings consisting of three different segmentation methods and three different ways of realizing the RCA classifier. In Fig.~\ref{fig:main-scatter} we provide the scatter plots of real versus predicted DSC for all nine settings with 525 data points each.

\begin{figure*}
\centering
  \includegraphics[width=1.0\linewidth]{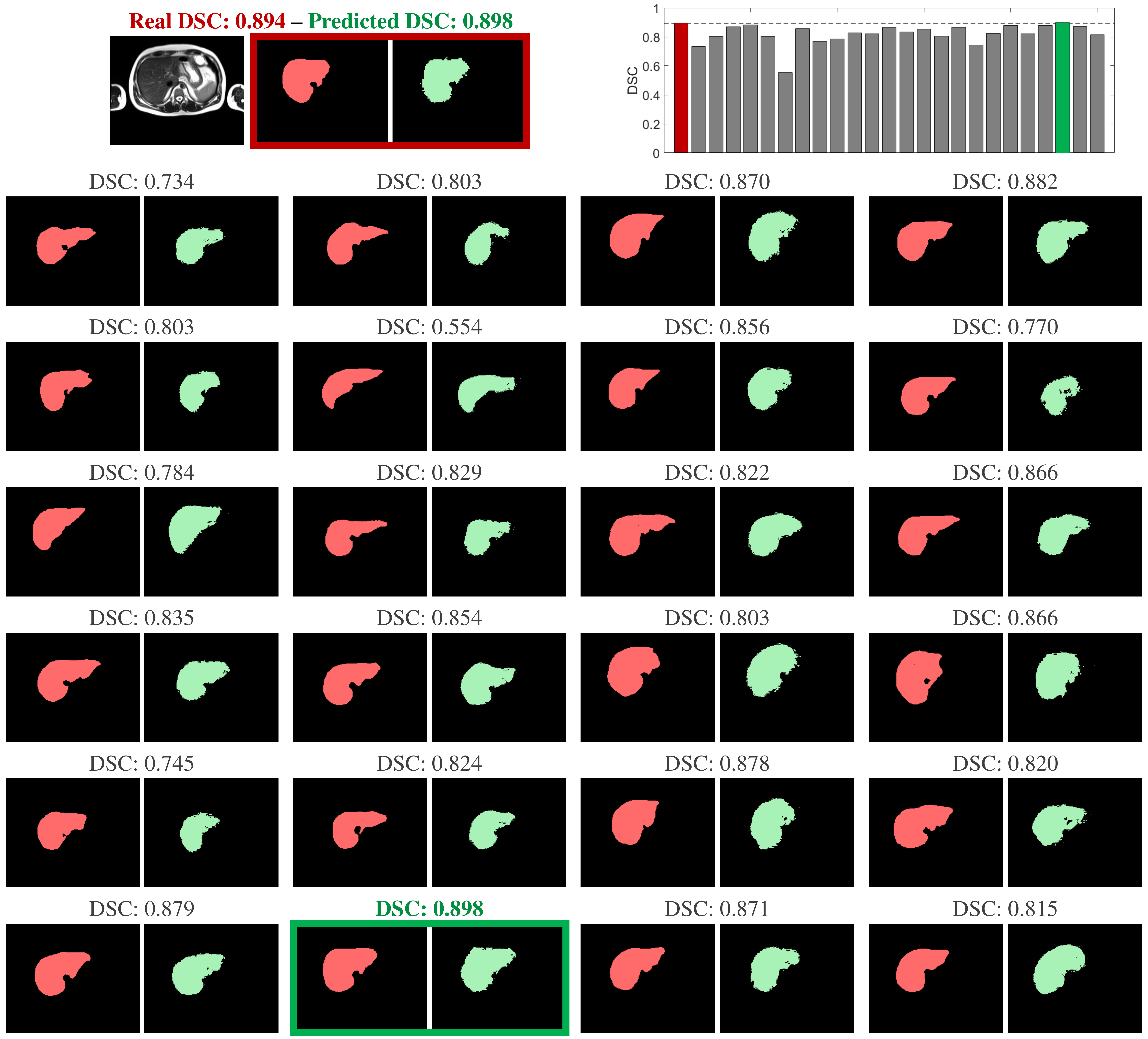}
\caption{Visual examples for predicting the segmentation quality of the liver for a new image shown on the top. Its GT segmentation (red colored) is unknown in practice, and we want to estimate the quality of the predicted, automatic segmentation shown on the most right (green colored). By taking the predicted segmentation as pseudo GT and training a RCA classifier we can obtain segmentations on a reference database with 24 images with available GT. The bar plot shows the real DSC in red and the different DSC values obtained for the reference images shown below. The green bar corresponds to the maximum DSC, and is selected as predicted DSC for the new image according to Eq.~(\ref{eq:rca}) which matches well the real DSC.}\label{fig:rca-high}
\end{figure*}

Overall, we observe high correlation between predicted and real DSC for both Atlas Forests and Single-Atlas when used as RCA classifiers, with the Single-Atlas showing correlations above 0.95 for all three segmentation methods. The Single-Atlas approach also yields the lowest MAEs between 0.05 and 0.07, and good 3-category classification accuracies between 81\% and 89\%. This is visually confirmed by the scatter plots in the right column of Fig.~\ref{fig:main-scatter} which show good linear relation close to the diagonal between predicted and real scores for most structures in the case where Random Forests or Multi-Atlas are used as the original segmentation method. When using Atlas Forests for RCA, we still observe good correlation but the relationship between predicted and real scores is off-diagonal with larger spread towards lower quality segmentation. The correlation is still good and above 0.82, MAEs are between 0.12 and 0.17 with classification accuracy going down to 0.62\%, 0.75\% and 0.78\% depending on the original segmentation method. For the case of the constrained CNNs, we observe that the prediction quality is lowest confirmed by the scatter plots and all quantitative measures, with correlations below 0.78 and MAEs above 0.2. The constrained CNNs seem to only work for predicting segmentation accuracy in case of major organs such as liver, lungs, and the spine but clearly struggle with smaller structures leading to many zero predictions even when the real DSC is rather high. This is most likely caused by the difficulty of training the CNNs with single images and small structures which does not provide sufficient amounts of training data.

\begin{figure*}[t]
\centering
	\includegraphics[width=0.49\linewidth]{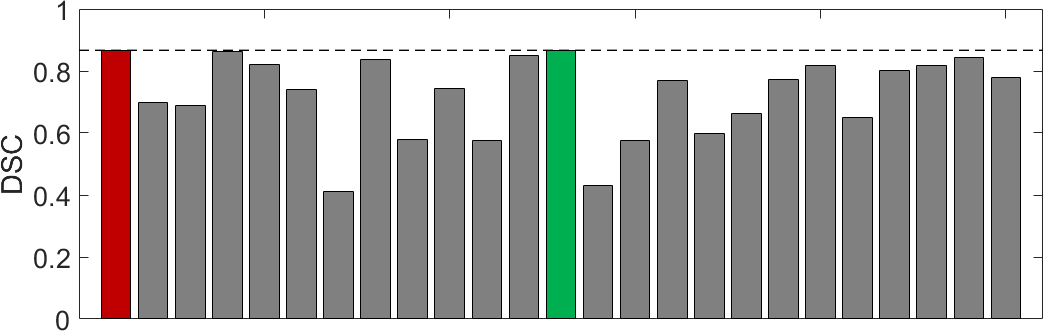}
	\includegraphics[width=0.49\linewidth]{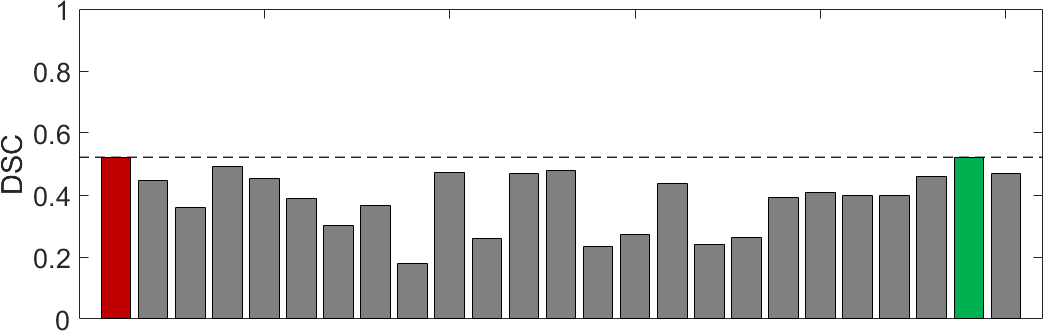}
\caption{The bar plots show two examples for predicting the real DSC (red) in case of a good quality (left) and bad quality segmentation (right) using a database with 24 reference images with available GT. The predicted DSC (green bar) selected according to Eq.~(\ref{eq:rca}) matches well the real DSC.}\label{fig:rca-examples}
\end{figure*}

Figure~\ref{fig:rca-high} shows an example for predicting the accuracy of a liver segmentation. Next to a slice from a T2w MRI volume we show the GT manual segmentation together with the result from a Random Forest. Underneath, we show the 24 segmentations obtained on the reference database when using the Single-Atlas RCA approach. The bar plot in the same figure shows the variation of the 24 DSC scores. Similarly, the bar plots in Fig.~\ref{fig:rca-examples} of two more examples illustrate the distribution of DSC scores when predicting a good quality segmentation on the left, and a poor quality segmentation on the right. The three examples support the hypothesis that selecting the maximum score across the reference database according to Eq.~(\ref{eq:rca}) is a good proxy for predicting segmentation quality.

Some of the original segmentation methods have problems segmenting structures such as the adrenal gland and clavicles. The CNNs, in particular, failed to segment adrenal glands in most cases. Because the real DSC for these is zero with no voxels labelled in the segmentation map, the RCA predictions are always correct as there are no labels for the RCA classifier for this structure. In order to investigate the effect of those zero predictions on the quantitative results, we also report in Tab.~\ref{tab:main-results} under the columns `No Zeros' the correlations, MAEs and classification accuracies when structures with a real DSC of zero are excluded. We observe that the zero predictions have mostly an impact on CNNs, either employed as original segmentation method or as RCA classifier. For Atlas Forests and Single-Atlas the effect on the accuracies is very little, confirming that those both are well suited within the RCA framework, independent of the original segmentation method.

\subsection{Detecting Segmentation Failure}
\begin{figure*}[t]
	\includegraphics[width=0.29\linewidth]{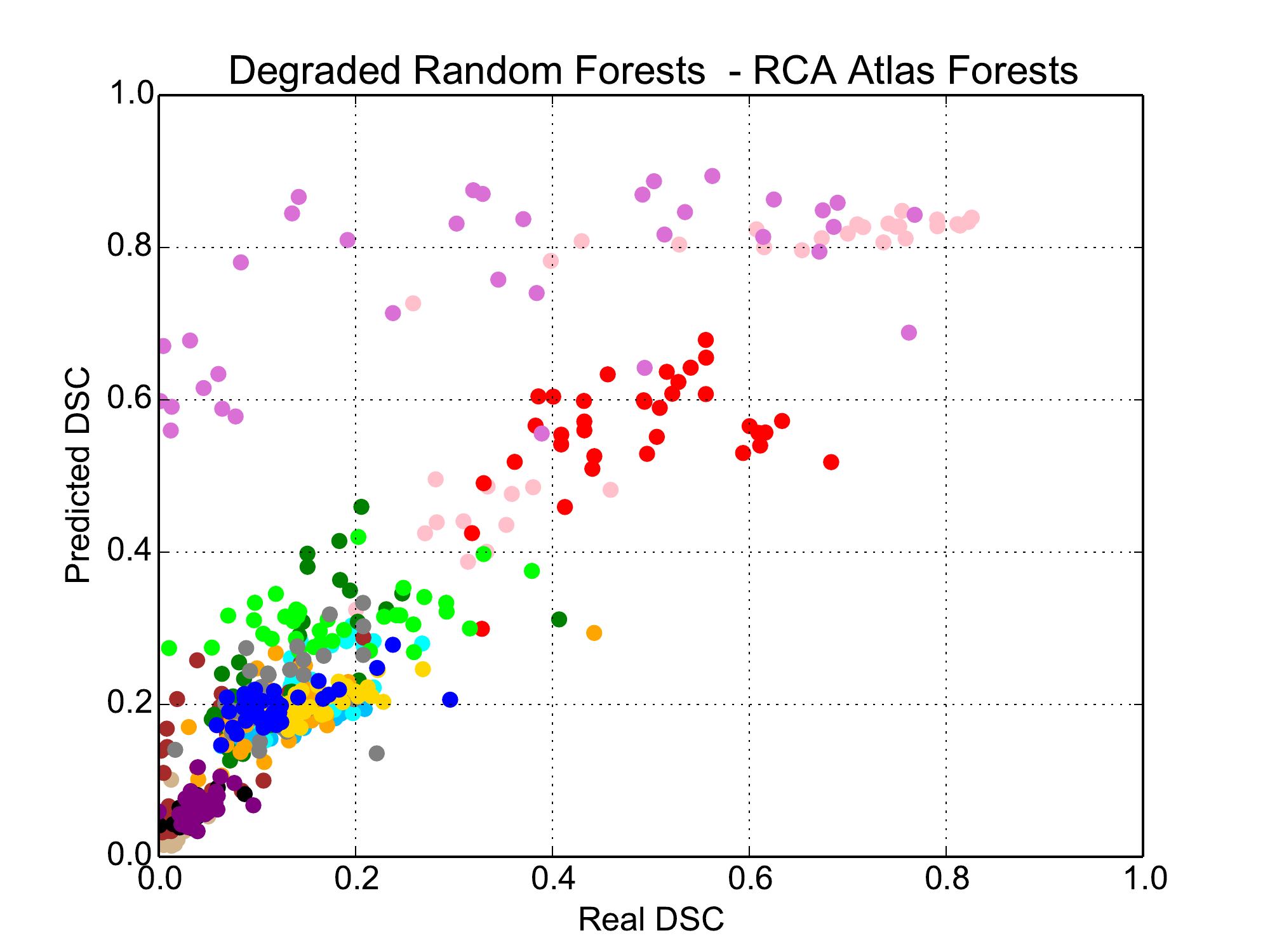}
	\includegraphics[width=0.29\linewidth]{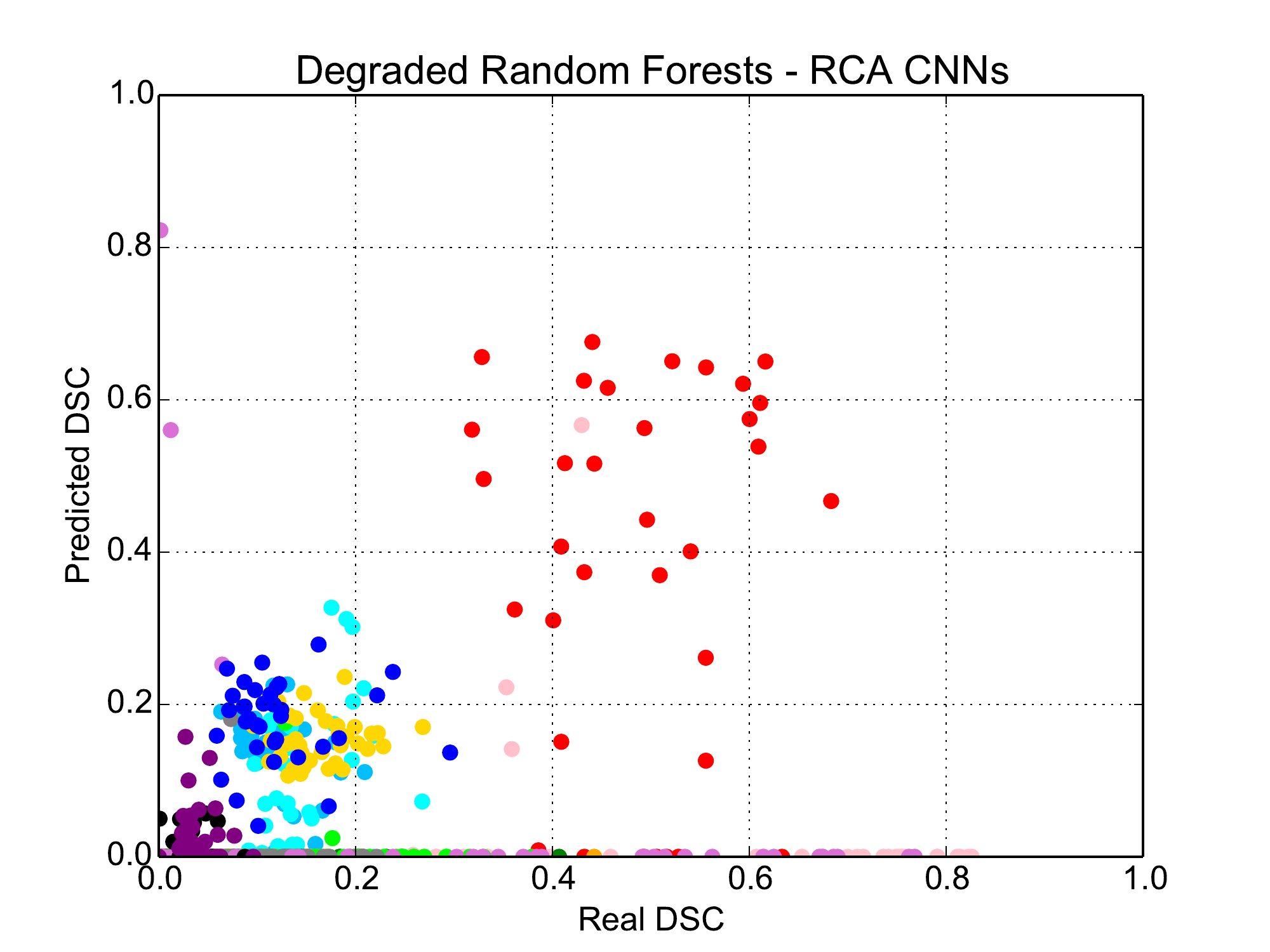}
	\includegraphics[width=0.29\linewidth]{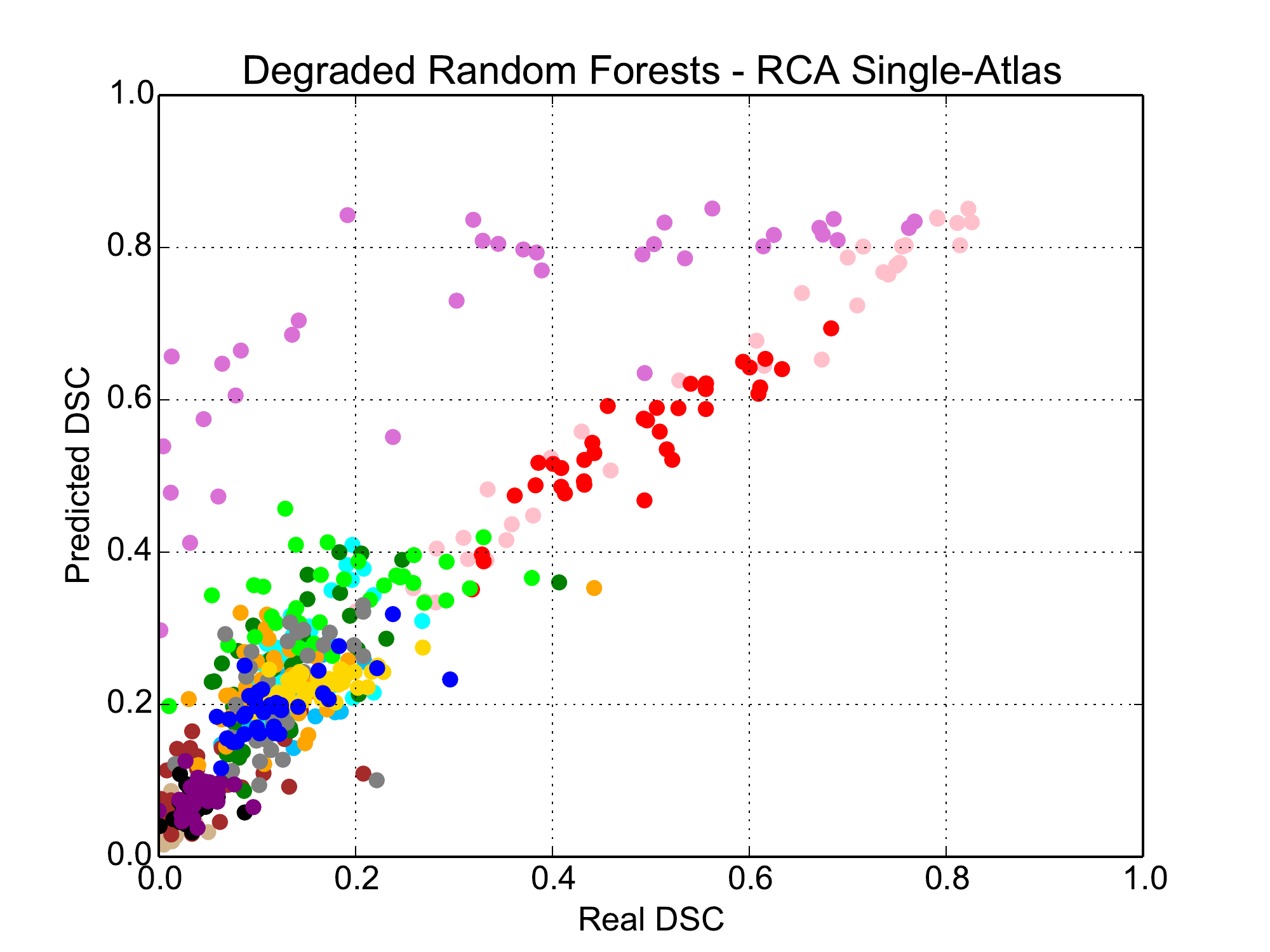}
	\includegraphics[width=0.10\linewidth]{artwork/legend.pdf}
\caption{Scatter plots for the experiment for detecting segmentation failure when using degraded Random Forests with limited depth as the segmentation method. Atlas Forests (left) and single-atlas label propagation (right) make highly accurate predictions in the low DSC ranges and thus, are able to correctly detect such failed segmentations, with the exception of the bladder. Constrained CNNs are again less suitable for RCA with many zero predictions.}\label{fig:degraded}
\end{figure*}

In clinical routine it is of great importance to be able to detect when an automated method fails. We conducted a dedicated experiment to investigate how well RCA can predict segmentation failure. From the scatter plots in Fig.~\ref{fig:main-scatter} we can see that all three segmentation methods perform reasonably well on most major organs with no failure cases among structures such as liver, heart, and lungs. In order to further demonstrate that RCA can predict failure cases in these structures, we utilize degraded Random Forests by limiting the tree depth at test time to 8. This leads to much worse segmentation results for most structures which is confirmed in the corresponding scatter plots shown in Fig.~\ref{fig:degraded}. Again, we evaluate the performance of the three different RCA classifiers, Atlas Forests, constrained CNNs and Single-Atlas. The results are summarized in Tab.~\ref{tab:deg_RF}. The constrained CNNs are again suffering from many zero predictions and less suitable for making accurate predictions. Atlas Forests and Single-Atlas, however, result in high correlations, low MAEs and very good classification accuracies. Low real DSC scores are correctly predicted and failed segmentations are identified. The only exception here is the bladder. This might be explained by the unique appearance of the bladder in the multi-spectral MRI with hyper-intensities in the T2w image, and its largely varying shape between subjects. It appears that even a badly segmented bladder can be sufficient for the RCA classifier to learn its appearance and segment the bladder well on at least one of the reference images. Overall, the experiment suggests that RCA with Atlas Forests and Single-Atlas can be employed in automatic quality control, for example, in large-scale studies where it is important to be able to detect failed segmentations which should be excluded from subsequent analyses.

\begin{table}[t]
\centering
\caption{Detecting segmentation failure}
\label{tab:deg_RF}
\begin{tabular}{l | c | c | c |}
RCA Classifier & Correlation & MAE & Accuracy 3-Cat. \\
\hline
Atlas Forests  & 0.853 & 0.096 & 0.884 \\
Constrained CNNs & 0.237 & 0.139 & 0.924 \\
Single-Atlas     & 0.875 & 0.097 & 0.928
\end{tabular}
\end{table}

\subsection{Results for Predicting Different Segmentation Metrics}
We further explore the ability to predict other evaluation metrics than DSC. We consider the following metrics: Jaccard index (JI), precision (PR), recall (RE), average surface distance (ASD), Hausdorff distance (HD) and relative volume difference (RVD). For this experiment, we use Random Forests as segmentation method, and Atlas Forests for RCA. The results are summarized in Tab.~\ref{tab:metrics}.

Good correlation is obtained between predicted and real overlap based scores, with low MAEs, and high accuracies. Since Jaccard is directly related to DSC\footnote{$JI=DSC/(2-DSC)$.}, it is expected that the predictions are of similar quality. Prediction accuracy for precision is lower than for recall. The two metrics capture different parts of segmentation error; under-segmentation is not reflected in precision, while over-segmentation is not captured in recall\footnote{$DSC=2 \cdot PR \cdot RE / (PR + RE)$.}. Distance based errors are unbounded, so we define thresholds for HD and ASD, and errors above are clipped to the threshold value, which is set to 150mm for HD, and 10mm for ASD. This also allows us to define ranges for the error categorization. For HD, we use the ranges $[0, 10]$, $(10,60]$, and $(60, 150]$ for good, medium and bad quality segmentations. For ASD we divide the range into $[0, 2]$ for good, $(2,5]$ for medium, and $(5,10]$ for bad segmentation quality. Compared to overlap based metrics, the RCA predictions for HD and ASD are not convincing with low correlation, high MAE, and low classification accuracy. RVD is the ratio of the absolute difference between the reference and predicted segmentation volume and the reference volume. Perfect segmentation will result in a value of zero. As RVD is also unbounded, we use a threshold of one to indicate maximum error. The predictions for RVD are good, with high classification accuracy of 0.68\%, similar to the overlap based scores. In conclusion, it seems RCA works very well for overlap based measures and for RVD to some extent, while distance based metrics cannot be accurately predicted with the current setting and would require further investigation.

\begin{table}[t]
\centering
\caption{Predicting different segmentation metrics}
\label{tab:metrics}
\begin{tabular}{l | c | c | c |}
Metric & Correlation & MAE  & Accuracy 3-Cat. \\
\hline
DSC {[}0 - 1{]} & 0.881 & 0.120 & 0.783   \\
JI {[}0 - 1{]} & 0.899 & 0.110  & 0.749  \\
PR {[}0 - 1{]} & 0.572 & 0.242  & 0.638  \\  
RE {[}0 - 1{]} & 0.833  & 0.170  & 0.663 \\
HD {[}0-150{]} & 0.189 & 41.38 & 0.387  \\
ASD {[}0-10{]} & 0.022 & 4.120  & 0.344 \\
RVD {[}0-1{]} & 0.474 & 0.051 & 0.678 
\end{tabular}
\end{table}

\section{Discussion and Conclusion}
The experimental validation of the RCA framework has shown that it is indeed possible to accurately predict the quality of segmentations in the absence of GT, with some limitations. We have explored different methods for realising the RCA classifier and could demonstrate that Atlas Forests and in particular, Single-Atlas label propagation yield accurate predictions for different segmentation methods. As the RCA framework is generic, other methods can be considered and it might be necessary to select the most appropriate one for the application at hand. We have also experimented with a constrained CNN trained on single images, which only works well for major organs such as liver, lungs and spine. There might be other more appropriate architectures for the purpose of RCA, which will be explored as part of future work.

An appealing property of the proposed framework is that unlike the supervised methods in \cite{frounchi2011automating} and \cite{kohlberger2012evaluating} no training data is required that captures examples of good and bad segmentations. Instead, in RCA we simply rely on the availability of a reference database with available GT segmentations. The drawback, however, is that we assume a linear relationship between predicted and real scores which should be close to an identity mapping, something we only found in the case of using Single-Atlas label propagation (cf. right column of Fig.~\ref{fig:main-scatter}). In the case of off-diagonal correlation, as for example found for Atlas Forests, an extension to RCA could be considered where the predictions are calibrated. This, however, requires training data from which a regression function could be learned, similar to \cite{kohlberger2012evaluating}. In order to demonstrate the potential of such an approach, we perform a simple experiment on the data that we used for conducting the main evaluation. After obtaining all predicted DSC scores, we run a leave-one-subject-out validation where in each fold we use Random Forest regression to calibrate the predictions. The results are summarized in Tab.~\ref{tab:regression} where we compare the quantitative measures before and after calibration. Both the MAEs and classification accuracies improve significantly for the case of Atlas Forests and constrained CNNs. For Single-Atlas, however, the results remain similar due to the already close to identity relationship between predicted and real scores before calibration. Calibration, however, comes with a risk of overfitting as the method will learn the relationship on the available training data but might not generalize to new data.

In our experiments we have found that best predictions are obtained for overlap based measures such as DSC and Jaccard Index. Whether those measures are sufficient to fully capture segmentation quality is debatable. Still, DSC is the most widely considered measure and being able to accurately predict DSC in the absence of GT has high practical value. Besides being useful for clinical applications where the goal is to identify failed segmentations after deployment of a segmentation method, we see an important application of RCA in large-scale imaging studies and analyses. In settings where thousands of images are automatically processed for the purpose of deriving population statistics, it is not feasible to employ manual quality control with visually inspection of the segmentation results. Here, RCA can be an effective tool to automatically extract the subset of high quality segmentations which can be used for subsequent analysis. We are currently exploring this in the context of population imaging on the UK Biobank imaging data where image data of more than 10,000 subjects is available which will be subsequently increased to 100,000 over the next couple of years. The UK Biobank data will enable the discovery of imaging biomarkers that correlate with non-imaging information such as lifestyle, demographics, and medical records. In the context of such large scale analysis, automatic quality control is a necessity and we believe the RCA framework makes an important contribution in this emerging area of biomedical research. In future work, we will further explore the use of RCA for other image analysis and segmentation tasks. To facilitate the wide application of RCA and use by other researchers, the implementations of all employed methods are made publicly available on the website of the Biomedical Image Analysis group\footnote{\url{https://biomedia.doc.ic.ac.uk/software/}}.

\begin{table*}[t]
\centering
\caption{Comparison of predicting DSC with and without calibration via regression.}
\label{tab:regression}
\begin{tabular}{ l  l | c c | c c | c c |}
& & \multicolumn{2}{c|}{Correlation} & \multicolumn{2}{c|}{MAE} & \multicolumn{2}{c|}{Accuracy 3-Categories} \\
Segmentation Method & RCA Classifier & Direct & Calibrated & Direct & Calibrated & Direct & Calibrated \\
\hline
Random Forests              & Atlas Forests          & 0.881            & 0.833       & 0.120             & 0.105      & 0.783         & 0.809                       \\
CNNs                        & Atlas Forests          & 0.828            & 0.929       & 0.166             & 0.079      & 0.623         & 0.783                       \\
Multi-Atlas                 & Atlas Forests          & 0.863            & 0.939       & 0.168             & 0.065      & 0.749         & 0.745                       \\
\hline                                                                                                                 
Random Forests              & Constrained CNNs       & 0.721            & 0.826       & 0.252             & 0.104      & 0.653         & 0.768                       \\
CNNs                        & Constrained CNNs       & 0.756            & 0.961       & 0.225             & 0.049      & 0.592         & 0.830                        \\
Multi-Atlas                 & Constrained CNNs       & 0.773            & 0.874       & 0.209             & 0.100      & 0.693         & 0.787                       \\
\hline                                                                                                                 
Random Forests              & Single-Atlas           & 0.955            & 0.872       & 0.051             & 0.089      & 0.888         & 0.815                       \\
CNNs                        & Single-Atlas           & 0.973            & 0.967       & 0.052             & 0.048      & 0.811         & 0.811                       \\
Multi-Atlas                 & Single-Atlas           & 0.962            & 0.918       & 0.067             & 0.080      & 0.822         & 0.825                      
\end{tabular}
\end{table*}

\section*{Acknowledgment}
This work is supported by the National Institute for Health Research (EME Project: 13/122/01) and the EPSRC First Grant scheme (EP/N023668/1). V. Valindria is supported by the Indonesia Endowment for Education (LPDP)- Indonesian Presidential PhD Scholarship programme. K. Kamnitsas is supported by the Imperial College President's PhD Scholarship programme. We gratefully acknowledge the support of NVIDIA Corporation with the donation of two Titan X GPUs for our research. The authors would also like to thank Darko Zikic for inspiring discussions on early ideas that have led to this work.

\ifCLASSOPTIONcaptionsoff
  \newpage
\fi

\bibliographystyle{IEEEtran}
\bibliography{refs}

\begin{thebibliography}{10}
\providecommand{\url}[1]{#1}
\csname url@samestyle\endcsname
\providecommand{\newblock}{\relax}
\providecommand{\bibinfo}[2]{#2}
\providecommand{\BIBentrySTDinterwordspacing}{\spaceskip=0pt\relax}
\providecommand{\BIBentryALTinterwordstretchfactor}{4}
\providecommand{\BIBentryALTinterwordspacing}{\spaceskip=\fontdimen2\font plus
\BIBentryALTinterwordstretchfactor\fontdimen3\font minus
  \fontdimen4\font\relax}
\providecommand{\BIBforeignlanguage}[2]{{%
\expandafter\ifx\csname l@#1\endcsname\relax
\typeout{** WARNING: IEEEtran.bst: No hyphenation pattern has been}%
\typeout{** loaded for the language `#1'. Using the pattern for}%
\typeout{** the default language instead.}%
\else
\language=\csname l@#1\endcsname
\fi
#2}}
\providecommand{\BIBdecl}{\relax}
\BIBdecl

\bibitem{boykov2006graph}
Y.~Boykov and G.~Funka-Lea, ``Graph cuts and efficient {ND} image
  segmentation,'' \emph{International Journal of Computer Vision}, vol.~70,
  no.~2, pp. 109--131, 2006.

\bibitem{iglesias2015multi}
J.~E. Iglesias and M.~R. Sabuncu, ``Multi-atlas segmentation of biomedical
  images: a survey,'' \emph{Medical Image Analysis}, vol.~24, no.~1, pp.
  205--219, 2015.

\bibitem{heimann2009statistical}
T.~Heimann and H.-P. Meinzer, ``Statistical shape models for {3D} medical image
  segmentation: a review,'' \emph{Medical Image Analysis}, vol.~13, no.~4, pp.
  543--563, 2009.

\bibitem{geremia2013classification}
E.~Geremia, D.~Zikic, O.~Clatz, B.~Menze, B.~Glocker, E.~Konukoglu, J.~Shotton,
  O.~Thomas, S.~Price, T.~Das \emph{et~al.}, ``Classification forests for
  semantic segmentation of brain lesions in multi-channel {MRI},'' in
  \emph{Decision Forests for Computer Vision and Medical Image Analysis}.\hskip
  1em plus 0.5em minus 0.4em\relax Springer, 2013, pp. 245--260.

\bibitem{zou2004statistical}
K.~H. Zou, S.~K. Warfield, A.~Bharatha, C.~M. Tempany, M.~R. Kaus, S.~J. Haker,
  W.~M. Wells, F.~A. Jolesz, and R.~Kikinis, ``Statistical validation of image
  segmentation quality based on a spatial overlap index 1: Scientific
  reports,'' \emph{Academic Radiology}, vol.~11, no.~2, pp. 178--189, 2004.

\bibitem{dice1945measures}
L.~R. Dice, ``Measures of the amount of ecologic association between species,''
  \emph{Ecology}, vol.~26, no.~3, pp. 297--302, 1945.

\bibitem{crum2006generalized}
W.~R. Crum, O.~Camara, and D.~L. Hill, ``Generalized overlap measures for
  evaluation and validation in medical image analysis,'' \emph{IEEE
  Transactions on Medical Imaging}, vol.~25, no.~11, pp. 1451--1461, 2006.

\bibitem{taha2015metrics}
A.~A. Taha and A.~Hanbury, ``Metrics for evaluating {3D} medical image
  segmentation: analysis, selection, and tool,'' \emph{BMC Medical Imaging},
  vol.~15, no.~1, p.~29, 2015.

\bibitem{deng2007simulating}
X.~Deng, L.~Zhu, Y.~Sun, C.~Xu, L.~Song, J.~Chen, R.~D. Merges, M.-P. Jolly,
  M.~Suehling, and X.~Xu, ``On simulating subjective evaluation using combined
  objective metrics for validation of {3D} tumor segmentation,'' in
  \emph{International Conference on Medical Image Computing and
  Computer-Assisted Intervention}.\hskip 1em plus 0.5em minus 0.4em\relax
  Springer, 2007, pp. 977--984.

\bibitem{ledig2014patch}
C.~Ledig, W.~Shi, W.~Bai, and D.~Rueckert, ``Patch-based evaluation of image
  segmentation,'' in \emph{Proceedings of the IEEE Conference on Computer
  Vision and Pattern Recognition}, 2014, pp. 3065--3072.

\bibitem{konukoglu2012discriminative}
E.~Konukoglu, B.~Glocker, D.~H. Ye, A.~Criminisi, and K.~M. Pohl,
  ``Discriminative segmentation-based evaluation through shape dissimilarity,''
  \emph{IEEE Transactions on Medical Imaging}, vol.~31, no.~12, pp. 2278--2289,
  2012.

\bibitem{van20073d}
B.~Van~Ginneken, T.~Heimann, and M.~Styner, ``3d segmentation in the clinic: A
  grand challenge,'' \emph{Proceedings of MICCAI Workshop on 3D Segmentation in
  the Clinic}, pp. 7--15, 2007.

\bibitem{sudlow2015uk}
C.~Sudlow, J.~Gallacher, N.~Allen, V.~Beral, P.~Burton, J.~Danesh, P.~Downey,
  P.~Elliott, J.~Green, M.~Landray \emph{et~al.}, ``{UK Biobank}: an open
  access resource for identifying the causes of a wide range of complex
  diseases of middle and old age,'' \emph{PLoS Med}, vol.~12, no.~3, p.
  e1001779, 2015.

\bibitem{dwork2015reusable}
C.~Dwork, V.~Feldman, M.~Hardt, T.~Pitassi, O.~Reingold, and A.~Roth, ``The
  reusable holdout: Preserving validity in adaptive data analysis,''
  \emph{Science}, vol. 349, no. 6248, pp. 636--638, 2015.

\bibitem{baraldi2005quality}
A.~Baraldi, L.~Bruzzone, and P.~Blonda, ``Quality assessment of classification
  and cluster maps without ground truth knowledge,'' \emph{IEEE Transactions on
  Geoscience and Remote Sensing}, vol.~43, no.~4, pp. 857--873, 2005.

\bibitem{liu2013no}
Y.~Liu, J.~Wang, S.~Cho, A.~Finkelstein, and S.~Rusinkiewicz, ``A no-reference
  metric for evaluating the quality of motion deblurring,'' \emph{ACM Trans.
  Graph.}, vol.~32, no.~6, pp. 175--1, 2013.

\bibitem{cerrato2011classification}
D.~Cerrato, R.~Jones, and A.~Gupta, ``Classification of proxy labeled examples
  for marketing segment generation,'' in \emph{Proceedings of the 17th ACM
  SIGKDD International Conference on Knowledge Discovery and Data
  Mining}.\hskip 1em plus 0.5em minus 0.4em\relax ACM, 2011, pp. 343--350.

\bibitem{correia2002stand}
P.~L. Correia and F.~Pereira, ``Stand-alone objective segmentation quality
  evaluation,'' \emph{EURASIP Journal on Applied Signal Processing}, vol. 2002,
  no.~1, pp. 389--400, 2002.

\bibitem{ge2007new}
F.~Ge, S.~Wang, and T.~Liu, ``New benchmark for image segmentation
  evaluation,'' \emph{Journal of Electronic Imaging}, vol.~16, no.~3, pp.
  033\,011--033\,011, 2007.

\bibitem{goldmann2008towards}
L.~Goldmann, T.~Adamek, P.~Vajda, M.~Karaman, R.~M{\"o}rzinger, E.~Galmar,
  T.~Sikora, N.~E. O?Connor, T.~Ha-Minh, T.~Ebrahimi \emph{et~al.}, ``Towards
  fully automatic image segmentation evaluation,'' in \emph{International
  Conference on Advanced Concepts for Intelligent Vision Systems}.\hskip 1em
  plus 0.5em minus 0.4em\relax Springer, 2008, pp. 566--577.

\bibitem{li2013benchmark}
H.~Li, J.~Cai, T.~N.~A. Nguyen, and J.~Zheng, ``A benchmark for semantic image
  segmentation,'' in \emph{2013 IEEE International Conference on Multimedia and
  Expo}.\hskip 1em plus 0.5em minus 0.4em\relax IEEE, 2013, pp. 1--6.

\bibitem{lamiroy2013computing}
B.~Lamiroy and T.~Sun, ``Computing precision and recall with missing or
  uncertain ground truth,'' in \emph{Graphics Recognition. New Trends and
  Challenges}.\hskip 1em plus 0.5em minus 0.4em\relax Springer, 2013, pp.
  149--162.

\bibitem{zhang2006meta}
H.~Zhang, S.~Cholleti, S.~A. Goldman, and J.~E. Fritts, ``Meta-evaluation of
  image segmentation using machine learning,'' in \emph{Computer Vision and
  Pattern Recognition (CVPR)}, vol.~1.\hskip 1em plus 0.5em minus 0.4em\relax
  IEEE, 2006, pp. 1138--1145.

\bibitem{chabrier2006unsupervised}
S.~Chabrier, B.~Emile, C.~Rosenberger, and H.~Laurent, ``Unsupervised
  performance evaluation of image segmentation,'' \emph{EURASIP Journal on
  Applied Signal Processing}, vol. 2006, pp. 217--217, 2006.

\bibitem{zhang2008image}
H.~Zhang, J.~E. Fritts, and S.~A. Goldman, ``Image segmentation evaluation: A
  survey of unsupervised methods,'' \emph{Computer Vision and Image
  Understanding}, vol. 110, no.~2, pp. 260--280, 2008.

\bibitem{unnikrishnan2007toward}
R.~Unnikrishnan, C.~Pantofaru, and M.~Hebert, ``Toward objective evaluation of
  image segmentation algorithms,'' \emph{IEEE Transactions on Pattern Analysis
  and Machine Intelligence}, vol.~29, no.~6, pp. 929--944, 2007.

\bibitem{warfield2004simultaneous}
S.~K. Warfield, K.~H. Zou, and W.~M. Wells, ``Simultaneous truth and
  performance level estimation {(STAPLE)}: an algorithm for the validation of
  image segmentation,'' \emph{IEEE Transactions on Medical Imaging}, vol.~23,
  no.~7, pp. 903--921, 2004.

\bibitem{li2010estimating}
X.~Li, B.~Aldridge, J.~Rees, and R.~Fisher, ``Estimating the ground truth from
  multiple individual segmentations with application to skin lesion
  segmentation,'' in \emph{Proc. Medical Image Understanding and Analysis
  Conference}, 2010, pp. 101--106.

\bibitem{bouix2007evaluating}
S.~Bouix, M.~Martin-Fernandez, L.~Ungar, M.~Nakamura, M.-S. Koo, R.~W.
  McCarley, and M.~E. Shenton, ``On evaluating brain tissue classifiers without
  a ground truth,'' \emph{NeuroImage}, vol.~36, no.~4, pp. 1207--1224, 2007.

\bibitem{sikka2010comparison}
K.~Sikka and T.~M. Deserno, ``Comparison of algorithms for ultrasound image
  segmentation without ground truth,'' in \emph{SPIE Medical Imaging}.\hskip
  1em plus 0.5em minus 0.4em\relax International Society for Optics and
  Photonics, 2010, pp. 76\,271C--76\,271C.

\bibitem{kohlberger2012evaluating}
T.~Kohlberger, V.~Singh, C.~Alvino, C.~Bahlmann, and L.~Grady, ``Evaluating
  segmentation error without ground truth,'' in \emph{International Conference
  on Medical Image Computing and Computer-Assisted Intervention}.\hskip 1em
  plus 0.5em minus 0.4em\relax Springer, 2012, pp. 528--536.

\bibitem{grady2012automatic}
L.~Grady, V.~Singh, T.~Kohlberger, C.~Alvino, and C.~Bahlmann, ``Automatic
  segmentation of unknown objects, with application to baggage security,'' in
  \emph{European Conference on Computer Vision (ECCV)}.\hskip 1em plus 0.5em
  minus 0.4em\relax Springer, 2012, pp. 430--444.

\bibitem{frounchi2011automating}
K.~Frounchi, L.~C. Briand, L.~Grady, Y.~Labiche, and R.~Subramanyan,
  ``Automating image segmentation verification and validation by learning test
  oracles,'' \emph{Information and Software Technology}, vol.~53, no.~12, pp.
  1337--1348, 2011.

\bibitem{zhong2010cross}
E.~Zhong, W.~Fan, Q.~Yang, O.~Verscheure, and J.~Ren, ``Cross validation
  framework to choose amongst models and datasets for transfer learning,'' in
  \emph{Joint European Conference on Machine Learning and Knowledge Discovery
  in Databases}.\hskip 1em plus 0.5em minus 0.4em\relax Springer, 2010, pp.
  547--562.

\bibitem{fan2006reverse}
W.~Fan and I.~Davidson, ``Reverse testing: an efficient framework to select
  amongst classifiers under sample selection bias,'' in \emph{Proceedings of
  the 12th ACM SIGKDD International Conference on Knowledge Discovery and Data
  Mining}.\hskip 1em plus 0.5em minus 0.4em\relax ACM, 2006, pp. 147--156.

\bibitem{zikic2014encoding}
D.~Zikic, B.~Glocker, and A.~Criminisi, ``Encoding atlases by randomized
  classification forests for efficient multi-atlas label propagation,''
  \emph{Medical Image Analysis}, vol.~18, no.~8, pp. 1262--1273, 2014.

\bibitem{breiman2001random}
L.~Breiman, ``Random forests,'' \emph{Machine Learning}, vol.~45, no.~1, pp.
  5--32, 2001.

\bibitem{criminisi2012decision}
A.~Criminisi, J.~Shotton, and E.~Konukoglu, ``Decision forests: A unified
  framework for classification, regression, density estimation, manifold
  learning and semi-supervised learning,'' \emph{Foundations and
  Trends{\textregistered} in Computer Graphics and Vision}, vol.~7, no. 2--3,
  pp. 81--227, 2012.

\bibitem{kamnitsas2016efficient}
K.~Kamnitsas, C.~Ledig, V.~F. Newcombe, J.~P. Simpson, A.~D. Kane, D.~K. Menon,
  D.~Rueckert, and B.~Glocker, ``Efficient multi-scale {3D} {CNN} with fully
  connected {CRF} for accurate brain lesion segmentation,'' \emph{Medical Image
  Analysis}, 2016.

\bibitem{bai2013probabilistic}
W.~Bai, W.~Shi, D.~P. O'Regan, T.~Tong, H.~Wang, S.~Jamil-Copley, N.~S. Peters,
  and D.~Rueckert, ``A probabilistic patch-based label fusion model for
  multi-atlas segmentation with registration refinement: application to cardiac
  {MR} images,'' \emph{IEEE Transactions on Medical Imaging}, vol.~32, no.~7,
  pp. 1302--1315, 2013.

\end{thebibliography}

\end{document}